\documentclass[12pt]{article}
\usepackage{enumitem}
\usepackage{fullpage}
\usepackage[english]{babel}
\usepackage[utf8]{inputenc}
\usepackage[authoryear,comma]{natbib}
\usepackage{amsmath,multibib}
\usepackage{amssymb,amsthm,mathabx,mathtools}
\usepackage{mathrsfs}
\usepackage{graphicx}
\usepackage{url}
\usepackage{caption,subcaption}
\usepackage{booktabs}
\usepackage{multirow}
\usepackage{caption}
\usepackage{epsfig,float}
\usepackage{url} % not crucial - just used below for the URL
\usepackage[usenames,dvipsnames]{color} 
\usepackage{listings}
\usepackage{mathrsfs,bm} 

\usepackage{tikz}
\usepackage{graphicx}

\theoremstyle{plain}
\newtheorem{thm}{Theorem}

\newcommand{\bthm}{\begin{thm}}
\newcommand{\ethm}{\end{thm}}
\newcommand{\bpf}{\begin{proof}}
\newcommand{\epf}{\end{proof}}
\theoremstyle{definition}
\newtheorem{defn}{Definition}
\newtheorem{rem}{Remark}
\usepackage[margin=.82in,a4paper]{geometry}
\usepackage[compact]{titlesec}
\numberwithin{equation}{section}
\usepackage{deep-macro}
\usepackage{epigraph}
\usepackage{relsize}

\newcommand{\qDIV}{\operatorname{qDIV}}
\newcommand{\qPivot}{\operatorname{qPivot}}

\usepackage[most]{tcolorbox}
\usepackage{tikz}
\usetikzlibrary{shadows.blur}
\usetikzlibrary{shapes.symbols}
\usetikzlibrary{shapes.geometric, arrows}
\tikzstyle{fitem} = [rectangle, shade, rounded corners, minimum width=4cm, 
       minimum height=1cm,text centered, draw=black, 
       top color=blue!20,
       bottom color=blue!17,
       rounded corners=6pt,
       blur shadow={shadow blur steps=2}]
\tikzstyle{root} = [rectangle, rounded corners, minimum width=4cm, 
       minimum height=1cm,text centered, draw=black, 
       fill=blue!5,
       rounded corners=6pt,
       blur shadow={shadow blur steps=7}]
\tikzstyle{arrow} = [ultra thick,->,>=stealth, shorten >= 2pt, shorten <= 2pt]
\tikzstyle{block} = [rectangle, draw, fill=blue!20,text centered, 
    text width=6em, rounded corners, minimum height=1cm]
\tikzstyle{tri} = [regular polygon, regular polygon sides=3, draw, 
				fill=orange!20, text width=7em, text centered, rounded corners, 
                minimum height=3em]   
\tikzstyle{round} = [ellipse, draw, fill=blue!20, 
    text width=5em, text centered, minimum height=5em]

\usepackage{bbm}
\usepackage{tabularx}
\newcolumntype{Y}{>{\centering\arraybackslash}X}
\newcolumntype{f}{>{\centering\arraybackslash}X}
\newcolumntype{h}{>{\hsize=.5\hsize\centering\arraybackslash\extracolsep{.1em}}X}

\newcolumntype{C}[1]{>{\hsize=#1\hsize\centering\arraybackslash}X}

\usepackage{deep-macro}
\DeclareMathOperator*{\argmin}{arg\,min}

\begin{document}
\begin{center} 
{\Large {\bf Breiman's ``Two Cultures'' Revisited and Reconciled}} 
\end{center}
\begin{center}
\begin{tabularx}{.97\linewidth}{ff}
{\bf Subhadeep Mukhopadhyay\footnote{The research in this paper was motivated by a conversation with Jerry Friedman, to whom I'm are very grateful. We have also benefited from discussions with Brad  Efron.
}} & {\bf Kaijun Wang}\\
\texttt{deep@unitedstatalgo.com} & \texttt{kwang2@fredhutch.org}
\end{tabularx}
\end{center}
%This research is funded by \texttt{h20.ai}.
%To a large extent
\begin{abstract} \nocite{breiman01}
In a landmark paper published in 2001, Leo Breiman described the tense standoff between two cultures of data modeling: parametric statistical and algorithmic machine learning. The cultural division between these two statistical learning frameworks has been growing at a steady pace in recent years.  What is the way forward? It has become blatantly obvious that this widening gap between `the two cultures' cannot be averted unless we find a way to blend them into a coherent whole. This article presents a solution by  establishing a link between the two cultures. Through examples, we describe the challenges and potential gains of this new \textit{integrated} statistical thinking.  
%to the fallacious ``either-or'' attitude 
%n apparently obvious 
%an attempt is made to take a `crude look at the whole' by establish a link between the two cultures.
%This article discusses one such approach in the context of regression. %However, more research is needed to better understand the challenges and potential gains from uniting data science front.
\end{abstract}
\vspace{-.425em}
\vskip.8em
\renewcommand{\baselinestretch}{.4}
\setlength{\parskip}{.2ex}
{\small
\setcounter{tocdepth}{2}
\tableofcontents
%\pagenumbering{gobble}
}
\vskip.5em
\linespread{1.28}
\renewcommand{\baselinestretch}{1.34}
\setlength{\parskip}{1.4ex}

\newpage
%%%%%%%%%%%%%%%%%%%%%%%%%%%%%%
\section{Introduction}
\vspace{-.6em}
\subsection{A Clash of Two Cultures}
\begin{center}
~~\textit{Some time around the year 2000 a split opened up in the world of statistics.}~\citep{efron2020pred}
\vspace{-.5em}
\end{center}
By the dawn of the 21st century, the statistics community was clearly divided into two distinct camps--namely, the (parametric) statistics and the (algorithmic) machine learning. Leo \cite{breiman01} presents a vivid description of this cultural polarization in his influential essay on ``Statistical Modeling: The Two Cultures.'' But what has triggered this split? 
\vskip.05em %In Breiman's book  Why should parametric statistical methods are less highly regarded than theoretical ones?
~~\texttt{ML$>$STAT}: \cite{breiman01} argued that algorithmic models are far more flexible, scalable, and accurate for complex big data problems. The traditional statistical methods, by contrast, are based on \textit{a priori assumed} parametric models that are mainly suitable for small datasets. 

%Why nor parametric models: Algorithmic modeling, large complex data sets and as a more accurate and informative alternative to Parametric modeling on smaller data sets. highly flexible nature and scalability for big data problems. parametric models are convince model..theory irrelevant..Led to irrelevant theory; Prevented statisticians from working on exciting
%new problems; not suitable for analysis by simple parametric models. `imposes an a priori straight jacket that restricts the
%ability of statisticians to deal with a wide range of statistical problems'
\vskip.05em
~~\texttt{STAT$>$ML}: This algorithmic-supremacy standpoint had been furiously debated by several eminent statisticians in the discussion of the original paper. They found Breiman's claim problematic on several counts. First, despite convincing prediction results, the absence of an explicit data-generating model (a probabilistic base) can make algorithmic methods less useful for scientific investigation; see the commentaries by \cite{cox01} and \cite{parzen01}.  \cite{efron2020pred} articulated this admirably: ``Abandoning mathematical models comes close to abandoning the historic scientific goal of understanding nature.'' Second, the black-box nature of algorithmic models makes them inscrutable, opaque, and not easily interpretable. In fact, regarding machine learning models, Breiman himself agreed that ``their mechanism for producing a prediction is difficult to understand...So on interpretability, they rate an F.''
\vskip.05em %The practical concern is that ..This extreme polarization is a  distributing reality, which has created an unbridgeable gulf between the two cultures.  as a realistic way forward
Where do we stand now?  To a large extent, the cultural division between these two statistical learning frameworks has been growing steadily over the past decades, especially in the post-deep-learning era. What options are we left with? 1) Let's get rid of black-boxes and use statistical parametric models, even though that means sacrificing accuracy; 2) Let's stick to black-boxes because they are accurate, even though that means sacrificing interpretability and robustness. Perhaps unsurprisingly, neither of these two extreme positions is tenable. To make real progress, we are forced to consider the question: how do these two different cultures fit together? Can we build a framework of data analysis that combines the best ideas from both sides and offer some way to make two into one? In this paper, we show that such an \textit{integrated} framework of statistical learning is entirely buildable. 

The message of this paper is very simple: the algorithmic models should not be feared because of their complexity, parametric models should not be shunned because of their simplicity, we can \textit{connect} them both to create a more powerful learning scheme. With the help of a synthetic example, the next section explains what we mean by `powerful' and why we need such a technology.

\subsection{The Butterfly Prediction Problem}
\begin{figure}[h]
 \centering
\includegraphics[width=.6\textwidth]{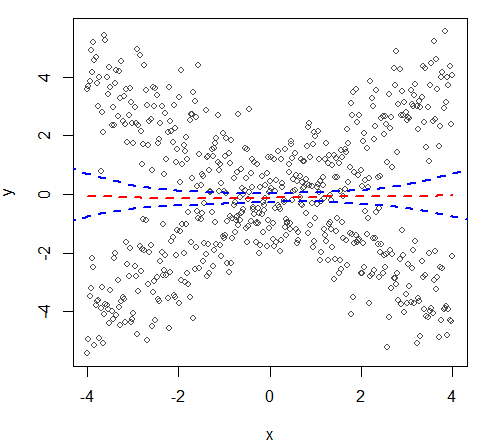}
\caption{The butterfly prediction problem: A superposition of $N=350$ samples, generated from $y=x+\epsilon$ and $y=-x+\epsilon$, with $\epsilon \sim \cN(0,1)$. The red line is the estimated regression function, whose 90\% bootstrap confidence band is shown by the blue dotted line.} \label{fig:butterfs}
\vspace{-.6em}
\end{figure}

\begin{quote} %This simple "butterfly" example is good enough to set the alarm bell ringing..
\textit{I strongly support the view that statisticians must face the crisis of the difficulties in their practice of regression.}~ \citep{parzen01}
\end{quote}
We are given $(x_i,y_i)$, $i=1,\ldots,N$ as shown in the Fig \ref{fig:butterfs}, which we refer to the \texttt{butterfly} data. The goal is to build a prediction model for the outcome variable $Y$ given $X=x$.

The conventional regression-based prediction methods proceed as follows: %by fitting curve to data usually 
\vskip.25em
~~(1) Estimation: We have a vast number of choices available. On the one hand, we have the simplest parametric model--the linear regression--and on the other, a large collection of sophisticated machine learning (ML) methods like k-nearest neighbors (knn), random forest, support-vector machine (svm), gradient boosting machine (gbm), neural net, etc. Irrespective of whether we choose a parametric or algorithmic method, the estimated-regression function takes the form, more or less, of a `flat' zero line, as shown in Fig. \ref{fig:butterfs}. 
\vskip.2em
~~(2) Inference: The obvious next question is, how accurate is the estimate? The blue dotted line in Fig. \ref{fig:butterfs} displays the 90\% bootstrap confidence band, which is narrow at the center (around $0$) and gets wider as we move outward--correctly reflecting the uncertainty of the mean function. Conclusion: the small bootstrap error (uncertainty) indicates that the estimated regression line is impressively accurate!
\vskip.25em %\footnote{Quantifying uncertainty of a prediction is an importance inferential task, but probably a more important task is to determine what is an appropriate \textit{estimand}.}
{\bf But, how useful is this analysis?} Even though we have an ``accurate'' estimate of the underlying regression prediction, it does not tell us anything useful. In fact, it paints a  grossly misleading picture of the reality (in the sense that $x$ is not useful for predicting $y$, which is clearly untrue). So, what went wrong? Do we need a bigger, more complex model? 
Better tuning of the hyperparameters? More computing power? More data? Unfortunately, `more data + more computing power' does not necessarily guarantee a better algorithm. The problem lies elsewhere.
%

%It is (unfortunately) easy to build a so-called efficient algorithm that does not serve any purpose. 
\vskip.1em
{\bf How to avoid statistical sin in the practice of regression?} 
To avoid  ``elegantly solving the wrong problem'' we must first ask: \textit{what} to estimate, before tackling the question of \textit{how} to estimate (linear or nonlinear; parametric or algorithmic; Bayes or frequentist). Almost all prediction methods start with the following `$\mu+\ep$' model\,\footnote{Open your favorite machine learning textbook, and it won’t be long before you encounter `$\mu+\ep$' model.}, which approximates $y$ as a non-linear function of $x=(x_1,\ldots,x_p)$:
\beq 
y~=~\mu(x)\,+\,\ep,
\eeq
where the random errors $\epsilon_i$'s are i.i.d with expected value zero, and finite variance; often assumed to be Gaussian with fixed variance $\ep \sim \cN(0,\si^2)$. The conventional statistical learning estimates the high-dimensional mean (regression) function $\mu(x)$, which can then be used for prediction. Now, it is not hard to see that no matter how complex the ML method we choose to approximate the function $\mu(x)$, it will still be fruitless (and a turbo-waste of computation\footnote{Algorithms are not bulldozer, they are tools for understanding.}) since the conditional mean contains no useful information for the \texttt{butterfly} example! But then, how should we perform prediction for the \texttt{butterfly} data? What should we report?  These are basic questions, yet often overlooked in the rush to apply fancy ML methods. 

%Stochastic prediction using a fixed number (here mean) is never a good idea.

%The autoML culture is based on the hypothesis that This suggests that improvements can arise from a higher computational budget and
%tuning more than fundamental algorithmic changes
\vskip.1em %changing the problem definition; rethink the problem of prediction..
{\bf Uncertainty prediction machine}. The main object of interest is the random variable $Y|X=x$, which will be  denoted by $Y_{|x}$. Can we really predict the value of the response variable $Y$ for a given $X=x$? No. It impossible to foresee what \textit{exact} value the random variable $Y_{|x}$ will take\footnote{After all, Heisenberg (1927) taught us, we can never know enough about the present to exactly predict the future event, except for predicting probabilities of different possible outcomes--the indeterminacy principle.}. At best, we can predict the probabilities and its distribution over all possible values of the response variable--the conditional distribution $f_{Y|X=x}(y)$, which encodes a \textit{complete} description of $Y_{|x}$. The problem of statistical learning is to design an `uncertainty prediction machine' to infer $f_{Y|X=x}(y)$ from the observed data $(x_i,y_i)$, $i=1,\ldots,N$.  Classical ML methods generally predict some type of summary statistic (e.g., location or scale) of the conditional distribution. However, the \texttt{butterfly} example taught us that this limited information could be dangerously misleading when dealing with messy heterogeneous datasets. One of the motivating questions behind our study is: how can we convert ``first generation'' machine learning methods into uncertainty distribution prediction machine? Our interest lies in extracting the underlying statistical ``law'' (governing equation) that is driving the data.

%Statistical prediction algorithms are not fortune-telling machine, they are probabilistic answer machine.

%Collapsing the distribution $f_{Y|X=x}(y)$ into summary statistic (e.g. location, scale etc.) is particularly dangerous Narrow prediction machine...they predicts ..Full-fledged uncertainty prediction machine.. 

%It seeks to upgrade {\bf "first generation" ML} methods, which predominantly act as a tool for ``fitting curve to data'' to ...upm that is interpretable and robust.   will be discussed next.

%%%%%%%%%%%%%%%%%%%%%%%

%%%%%%%%%%%%%%%%%%%%%%%%%%%%%%%%
\subsection{Uncertainty Prediction Machine: What to Expect?}
Before delving into the technical details, it may be worthwhile if we start by taking a `crude look at the whole' to clearly convey what practical results can be hoped from our statistical learning technology. Given the data $(x_1,y_1),\ldots,(x_N,y_N)$ with $x_i \in \cR^p$ and $y_i \in \cR$, our primary focus is the inference of $Y_{|x}$, where $Y_{|x}$ denotes the conditional random variable $Y|X=x$ with distribution function $F_{Y|X=x}(y)$ and density $f_{Y|X=x}(y)$. We will be using the \texttt{butterfly} data as a stylized example to describe the different stages of our analysis.

%are presented through
%a demonstrative example 

%Thus, a decrease in uncertainty corresponds to an increase in information.
\vskip.24em
%%%%%%%%%%%%%%%%%%%%%%%%%
{\bf Level 0}. \textit{ML curve-fitting}.  We start with a user-selected machine learning method ML$_0$ to estimate (possibly high-dimensional) regression function $\widehat{\mu}(x)$ along with its uncertainty band. The result for \texttt{butterfly} data is already displayed in Fig. \ref{fig:butterfs}. Frankly speaking, this is where the practice of traditional machine learning ends, and ours begins.

%can give a fairly interpretable picture of the relationship between $x$ and $y$  

%If not, then can we provide further insight into the \textit{nature} of the misfit? Finally, can we \textit{repair and refine} the starting ML method?

\vskip.24em
%%%%%%%%%%%%%%%%%%%%%%%%%
{\bf Level 1}. \textit{Heterogeneity diagnostic}. A natural question is whether the initial ML$_0$ captured the essential relationship between $y$ and the predictor variables. This can be checked by searching for (unmodeled) patterns in the residuals. Recall that for the \texttt{butterfly} data $\hat \mu(x)\approx 0$ for all $x$; thus residuals are actually the original response values $y_1,\ldots,y_n$. 

\begin{figure}
\begin{subfigure}[t]{.45\linewidth}
    \centering
    \caption{Heterogeneity Diagnostic Plot}
        \includegraphics[width=\linewidth,trim=1cm 0cm 0cm .5cm]{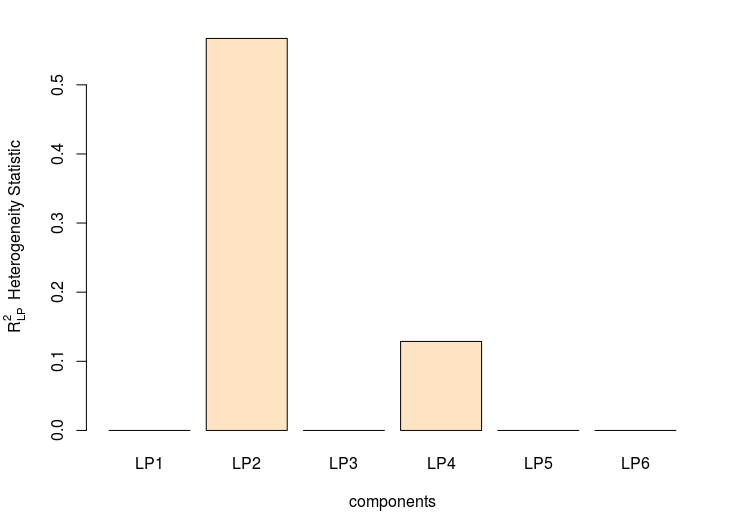}
    \end{subfigure}~~
    \begin{subfigure}[t]{.45\linewidth}
    \centering
    \caption{Pivot Density at X=2}
        \includegraphics[width=\linewidth,trim=1cm 0cm 0cm .5cm]{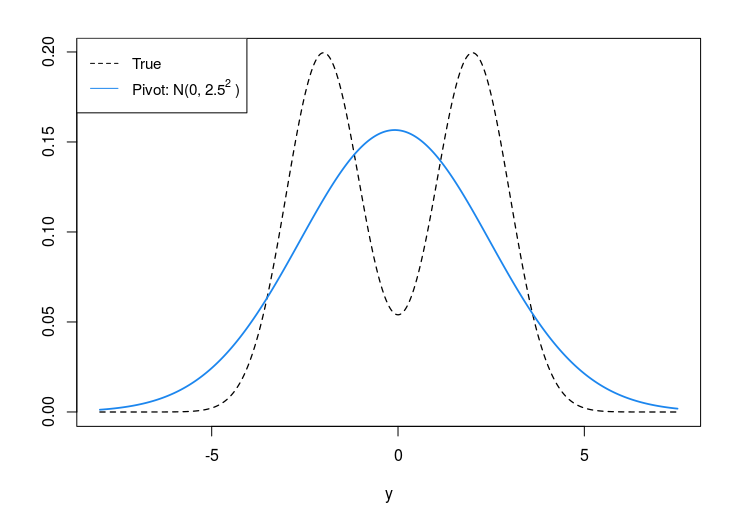}
    \end{subfigure}\\[1.5em]
    \begin{subfigure}[t]{.45\linewidth}
    \centering
    \caption{Uncertainty of Pivot Model}
        \includegraphics[width=.9\linewidth,trim=2cm 0cm 0cm .5cm]{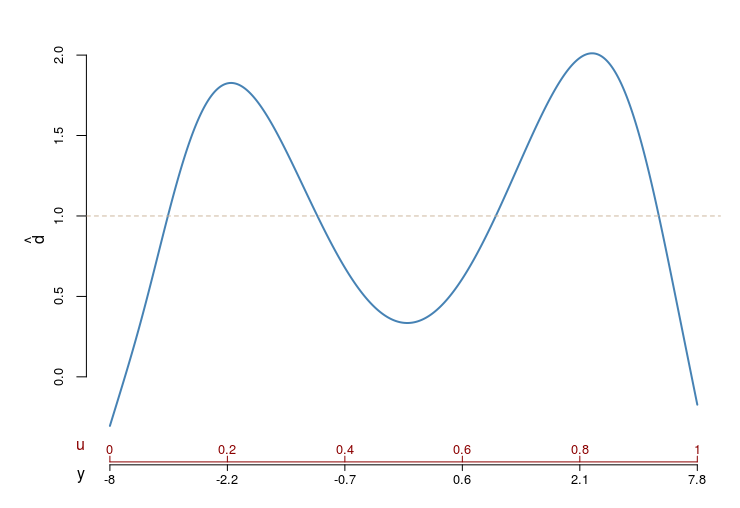}
    \end{subfigure}~~
    \begin{subfigure}[t]{.45\linewidth}
    \centering
    \caption{Predictive Inference}
         \includegraphics[width=\linewidth,trim=1cm 0cm 0cm .5cm]{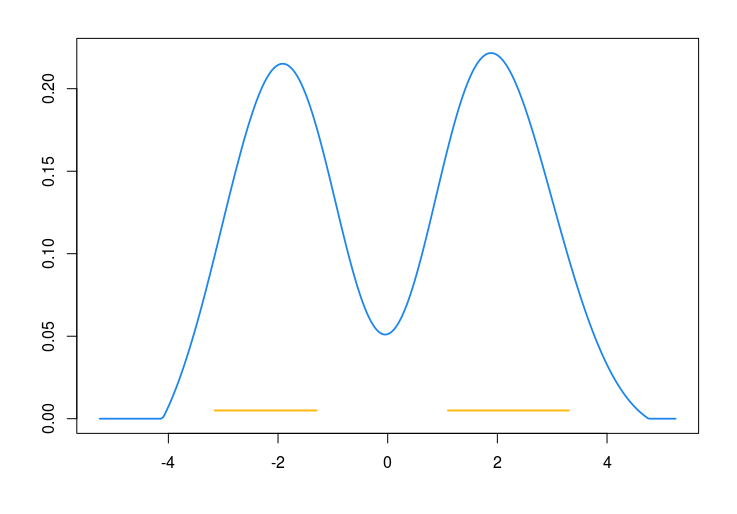}
    \end{subfigure}\\[1em]
    \begin{subfigure}[t]{.45\linewidth}
    \centering
    \caption{Quantile Prediction}
        \includegraphics[width=\linewidth,trim=1cm 0cm 0cm .5cm]{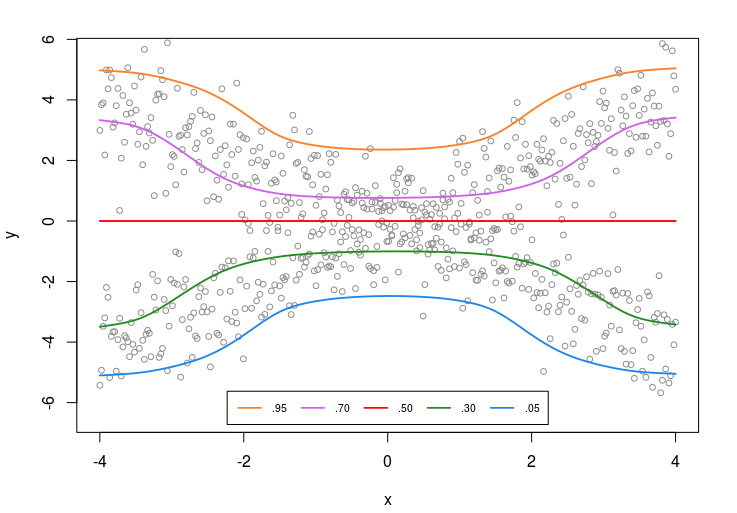}   
        \end{subfigure}~~
        \begin{subfigure}[t]{.45\linewidth}
        \centering
    \caption{Goodness-of-fit}
         \includegraphics[width=\linewidth,trim=1cm 0cm 0cm .5cm]{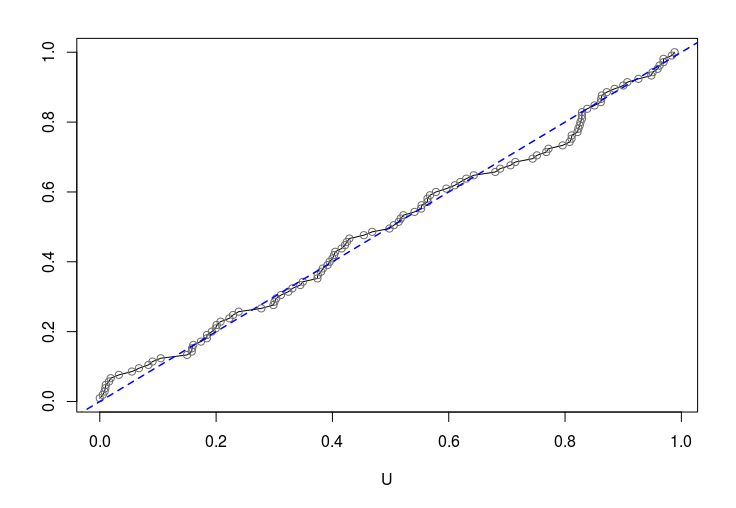}
    \end{subfigure}
    \caption{Analysis of Butterfly data:  (a) Heterogeneity component analysis for residuals. It decomposes total deviation from homogeneity into explainable components. Here, the starting ML$_0$ (which we have selected to be knn method with $k=15$) fails to capture the varying scale and tail. (b) Gaussian pivot $f_0(y)$ shown in blue; black dotted line denotes the true $\frac{1}{2}\cN(2,1)+\frac{1}{2}\cN(-2,1)$. (c) The estimated bimodal $\whd_{x_0}^0(u)$. (d) The shape of $d$-modulated density function Eq.\eqref{eq:beq}, which has an inbuilt `self-correcting' mechanism through $\whd_{x_0}^0$. (e) Estimated conditional quantile curves for $u=.05,.30,.50,.70$ and $.95$. (f) QQ-plot for checking the uniformity of generalized quantile-residuals $U_i=\whF_{Y|X=x_i}(y_i)$ on a 15\% hold-out set.} \label{fig:bfly}
        \end{figure}

The critical task boils down to checking whether the residuals are homogeneous, i.e., $f_{Y|X=x}(y)$ $=f_Y(y)$ for all $x$. This would ensure that ML$_0$ was able to capture all the important predictive information. However, in most practical real-life problems (including \texttt{butterfly} data) the residuals will be heterogeneous. And the main challenge is to identify the components of $f_{Y|X=x}(y)$ that are significantly changing with $x$.  We like to  address this by constructing a simple diagnostic plot that can help users to visually examine and detect the prominent \textit{sources} of heterogeneity in the residuals. Fig. \ref{fig:bfly}(a) is one such graphical exploratory plot drawn for the   
\texttt{butterfly} data, which reveals scale (second-order effect) and tail (fourth-order kurtosis effect) of $f_{Y|X=x}(y)$ are changing significantly with $x$. Accordingly, this tool helps us to \textit{discover} the hidden patterns that, while invisible to ML$_0$, still persist and should be modeled for sharpening the prediction quality.

%\texttt{RAPID} plot: {\bf \underline{R}esidu\underline{a}ls \underline{P}red\underline{i}ctability \underline{D}iagnostic}

%It seems intuitively clear that...

\vskip.24em

%%%%%%%%%%%%%%%%%%%%%%%%%
{\bf Level 2}. \textit{Pivot density}. We now
turn our attention to the problem of predicting conditional distribution at a particular $x$, say at $x_0=2$ for the \texttt{butterfly} example. There are some obvious candidates: $\wtf_Y(y)$ the empirical (marginal) distribution of $y$; or its nonparametric smooth (say, kernel-smoothed) version $\fhat_Y(y)$; or some kind of parametric model motivated from domain-knowledge or statistical convenience, e.g., $\cN(\hat \mu(x_0),\hat \si_y^2)$, where $\hat \si_y$ denotes the standard error of $y$. We call these initial starting ``guesses'' the pivot density; they are nothing but a crude first-approximation of the true \textit{unknown} conditional density estimate. For the \texttt{butterfly} data, we choose $\cN(0,\hat \si_y^2=2.55^2)$ to be the pivot, as shown in Fig. \ref{fig:bfly}(b). 

A natural question to ask is whether the presumed pivot model is approximately correct. In most practical cases, the answer will be no, they are not adequate. Thus we have to take some corrective measures to rectify the deficiency of the starting pivot. But before doing so, it is crucial to devise some technique that can reveal the deficiency of the current model. The only problem is that we don't have enough $y$-samples at $X=x_0$ to perform traditional goodness-of-fit tests. We have to come up with something \textit{new} that is computable and interpretable.

%viewed as a ``frequentist's prior'' without Bayes rule ..

%Lesson learned: It is (unfortunately) easy to build an efficient algorithm that does not serve any purpose. Start with a simple approximately correct model..

\vskip.24em
%%%%%%%%%%%%%%%%%%%%%%%%%
{\bf Level 3}. \textit{Uncertainty quantification for the pivot}.  The basic question is whether the starting pivot density $f_0(y)$ is a good model for $Y_{|x_0}$. We answer this question by estimating the following \textit{contrast} density function for $u=F_0(y)$
\beq \label{eq:du}
d(u;F_0,F_{Y|X=x}):=\,d_x^0(u)\,=\dfrac{f_{Y|X=x} ( F_0^{-1}(u))}{f_0( F_0^{-1}(u))},~~0<u<1\eeq
where $F_0^{-1}(u)$ is the quantile function for the pivot $f_0(y)$. Note that $d_x^0(u)$ captures the deviation of $f_0(y)$ from the true \textit{unknown} conditional density $f_{Y|X=x_0}(y)$, and by virtue of doing so it helps us to quantify and characterize the uncertainty of the pivot density. By inspecting the shape of $d_{x_0}^0(u), 0<u<1$ (how it deviates from uniformity), practitioners can quickly infer whether $f_0(y)$ needs any repair or not.

Fig. \ref{fig:bfly}(c) displays the estimated contrast function for the \texttt{butterfly} data. Looking at the shape, it becomes clear that the initial Gaussian pivot completely missed the bimodality of $f_{Y|X=2}(y)$. The contrast function $d_x^0(u)$ plays a central role in our theory. In Section \ref{sec:dest} we provide an estimation strategy. The fascinating aspect of the algorithm is that it utilizes the power of the initial machine learning algorithm ML$_0$ to get the covariate-adaptive $\whd_x^0$.

%This could be a valuable tool for 
\vskip.24em
%%%%%%%%%%%%%%%%%%%%%%%%%
{\bf Level 4.} \textit{Density modulation}.
When the contrast function indicates the inadequacy of the starting pivot $f_0(y)$ as a model for the conditional distribution $f_{Y|X=x}(y)$, what to do next? How to modify (or `boost') $f_0(y)$ to make it data-consistent? This can be solved in a remarkably simple way via \textit{density modulation}: $\hf_{Y|X=x}=f_0 \times \whd_x^0$, where $d_{x}^0$ modulates (modifies) the shape of starting $f_0(y)$ to produce the conditional density. The detailed theory of why and how this works will be discussed in Section 2.  Fig. \ref{fig:bfly}(d) displays the estimated predictive density for the \texttt{butterfly} data, which we can write down analytically as
\beq \label{eq:beq}
\hf_{Y|X=x}(y),=\,\dfrac{1}{\hat \si} \phi\big(\frac{y}{\hat \si}\big) \times \big\{1 -0.20\, T_2(y;F_0) - \,0.47\,T_4(y;F_0)\big\},~~ \text{where $\hat \si=2.55$}\eeq
which is product of two components: the Gaussian pivot and
the contrast density $d_{x_0}^0(F_0(y))$. The basis functions $\{T_j(y;F_0)\}$ in equation \eqref{eq:beq} are specially-designed rank-based orthonormal polynomials of the pivot. They injects robustness and  non-linearity into the data modeling process--but more on this later. We will also show that this new class of $d$-modulated conditional density models can generate an extraordinarily rich spectrum of shapes.

One notable aspect of \eqref{eq:beq} is that we have `identified' a parametric model (instead of blindly \textit{assuming}, which Breiman sharply criticized) by algorithmic (fully nonparametric) means. In Section \ref{sec:dest}, we elucidate the process by which we boost and convert the ``$0$th order'' machine learning method (ML$_0$) into a finitely parameterizable uncertainty prediction machine. This allows our learning philosophy to \textit{integrate} the fundamental character of both algorithmic data-driven modeling and parametric statistical modeling. Nevertheless, the predicted distribution \eqref{eq:beq} compactly represents all possible outcome values along with their respective probabilities, which is essential for decision-making under uncertainty.

%thereby capturing the ``intrinsic'' uncertainties for making informed decisions.

%requires the whole distribution of $Y_{|\xb}$. The  $f_{Y|X=x}(y)$ 

%surprising array of shapes ..This simple model can possess ..

%of the a priori selected ..exploratory graphical diagnostic and uncertainty quantification. comprehensively the uncertainty about the choice of 

%Condensed the ML model into [quote Brad efron no parameter)].....suddenly condensed in to few interpretable parameters..derives a succinct representation ..which is amiable for statistical inference.

%We call it ``coupling'' conditional density  model (\texttt{CCD}, in short).

%The conditional distribution for $X=x_0$ is translating $\hf_{Y|x_0}(y)$ by the amount $\hat\mu(x_0)$.  $\hf_{Y|x_0}(y-\hat\mu(x_0))$

%Insight into the underlying statistical ``law'' that is driving the data. 

\vskip.24em
{\bf Level 5}. \textit{Predictive inference}: What are some of the most likely values of the response variable $y$, given that we have observed $X=x_0$? For a fixed $\al$, ideally, we would like to find the smallest volume region that covers $100(1-\al)\%$ area of the conditional distribution, aka the highest density prediction region. The orange highlighted part in Fig. \ref{fig:bfly}(d) displays 68\% highest-density prediction region for the \texttt{butterfly} data, which consists of  two disjoint intervals [-3.33,-1.40] $\cup$  [1.20,3.28]. The volume of this set (the sum of the lengths of the intervals) can be used to quantify the \textit{uncertainty} of the prediction.  Additionally, we can extract all the conditional quantiles from the estimated $\whF_{Y|X=x}(y)$; see panel (e). Our quantile regression curves are, by design, non-crossing and nonlinear. Curious readers may contrast this with the traditional quantile regression estimator, as shown in Fig. \ref{fig:kqr} of Appendix A.4.1.

%Looking at the multi-modal shape of the conditional density in , it is obvious that mean nor median is a not good summary of the distribution. 

\vskip.24em
%%%%%%%%%%%%%%%%%%%%%%%%%
{\bf Level 6}. \textit{Verification}: Data modeling typically consists of four stages: estimation (of the initial pivot), exploration (discovering the deficiency of the pivot), rectification (repairing the pivot), and verification (checking whether the corrected pivot emulates the observed reality). The last step remains to be done, which requires \textit{evidence} that the estimated uncertainty prediction model provides a satisfactory description of the data. Scientific data analysis, by definition, needs to be testable (falsifiable), and there's no two ways about it.

%\footnote[2]{As Karl Popper said: ``if a theory is falsifiable, then it is scientific; if it is not, then it is not science.''}

For our \texttt{butterfly} example, we compute $U_i=\whF_{Y|X=x_i}(y_i)$ for $i=1,\ldots,N$ using our trained model on a 15\% hold-out validation dataset. It is not difficult to see that the distribution of $U_i$'s will be `close' to \texttt{Uniform}[0,1], if the estimated conditional density can explain the essential patterns/variations in the data. The quantiles of the empirical distribution of $U_i$'s are plotted against the respective quantiles of the uniform distribution at the bottom right of Fig. \ref{fig:bfly}. The tight cluster of points around the 45-degree reference line strongly indicates that the model is successful in capturing the underlying data generating process, and can be trusted to make predictions. Additionally, in Section \ref{sec:gof}, we provide a formal goodness-of-fit test for uniformity, which, when applied for this example, yields a p-value of $0.82$.

%\begin{rem}
%ML acts as a rectifier of the conventional statistical prediction algorithms. marriage of two cultures rather than a battle of two cultures. Where one complements the strength of other. Because of that we call it ``\textit{Integrative Statistical Learning}.'' by embracing algorithmic, parametric and exploratory philosophies. 
%\end{rem}

\begin{rem}
Before leaving this section, we make two comments: (1) The purpose of this \texttt{butterfly} data example was to explain how our data-analysis scheme systematically extracts increasingly fine-grained knowledge from data, starting from a user-specified ML$_0$. (2) Our statistical learning theory is simultaneously applicable to small and large-scale high-dimensional problems.  The next section clearly outlines our goals and objectives in somewhat broader terms.
\end{rem}

%Next section outlines the goals in clear terms

%that we want to achieve (at least approximately).

%described  as a chain of well connected steps that .

%guiding principles

\begin{figure}[t]
    \centering
     \includegraphics[width=\linewidth]{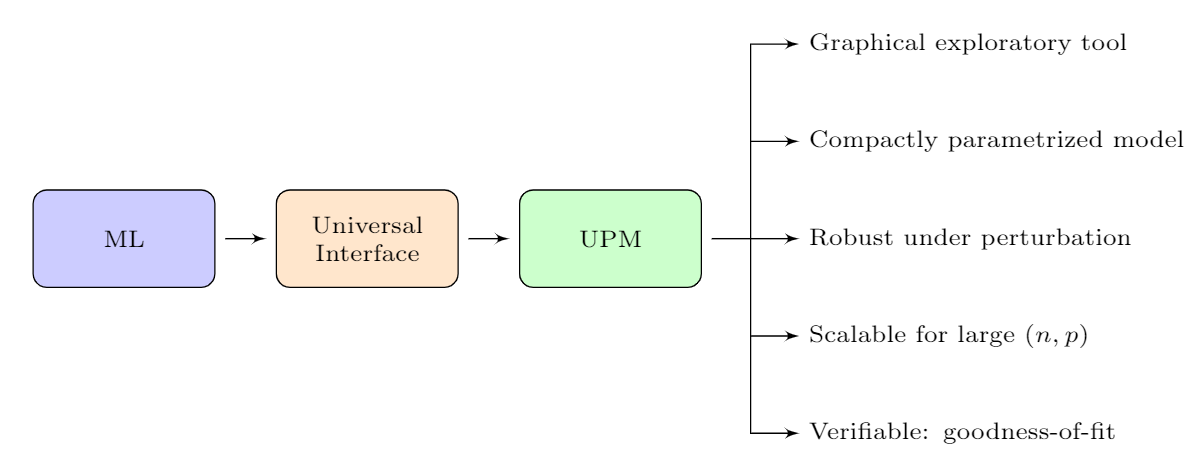}
     \vskip1em
    \caption{Integrated statistical learning framework at a glance; `ML' stands for (an arbitrary) machine learning algorithm, and `UPM' denotes uncertainty prediction machine.}
    \label{fig:upm_flowchart}
\end{figure}
%%%%%%%%%%%%%%%%%%%%%%%%%%%%%%%%%%%
\subsection{Goals and Contributions}
Figure \ref{fig:upm_flowchart} shows a schematic description of the modular architecture of our data analysis pipeline, which aims to ``assemble'' different statistical learning cultures--algorithmic, parametric, exploratory, and robust modeling--into one coherent whole.

%which attempts in this direction by establishing an important link between statistical and machine learning cultures, with the hope to facilitate better communication across the great divide. 

~~$\bullet$  Unified Interface:   We like to construct a generic ``interface'' that can convert \textit{any} algorithmic regression procedure into an uncertainty prediction machine. We are less interested in solutions that depend on the specific choice of an ML method; developing `tricks' on a case-by-case basis is too clumsy. This raises the question:  how to design such a simple and unified interface? Section \ref{sec:theory} presents some general theory that enables such a design. 
%strategy. 

\vskip.1em
~~$\bullet$ Algorithmic Parameterization:  The next crucial aspect of our method lies in distilling an explicit and simple statistical model from a black-box technique. Section \ref{sec:theory} lays out such a mechanism to allow algorithmic synthesis of probabilistic models with interpretable statistical parameters. This opens up the possibility of building closed-form, interpretable statistical models that are as expressive and scalable as machine learning models. 
\vskip.1em
%to combine the strength of algorithmic scalability with parametric interpretability.

~~$\bullet$ Exploratory Machine Learning: How can we check whether a trained machine learning method is consistent with the observed data? If not, can we discover its \textit{blind spots} using some exploratory graphical tools? Finally, can we revise the starting algorithmic method when required? All these basic questions are often left unanswered by traditional machine learning procedures.  In the words of Richard \cite{hamming97art}, ``AI has traditionally stuck to the \textit{what} is done and seldom considered the \textit{how} it is done.'' The `how' part is important to build trusted AI/ML systems. Section \ref{sec:xml} discusses some new tools to make the `modeling process' more transparent and interpretable.\footnote{
A recently-passed legislation of New York City Council says that  ``As we advance into the 21st century, we must ensure our government is not black boxed.''} It consists of the following chain of well-connected steps:
%chain of well connected steps
\vskip.1em
~~~Data + Pivot $\rightarrow$ Exploration $\rightarrow$ Discovery $\rightarrow$ Rectification $\rightarrow$ Verification  $\rightarrow$ Prediction. 
\vskip.05em
This layout of data analysis embodies the viewpoint that empirical (statistical) science is a  `self-correcting' process. 
 %which has  the following layout
%The purpose of these chain of well connected steps is to build trust in ML-models by making it , and falsifiable. 

\vskip.15em
~~$\bullet$ `Robust + Efficient' Machine Learning: Current ML systems are extremely vulnerable to small perturbations of the training data, which makes them ineffective for critical applications. One of the unsolved dilemmas is to construct learning algorithms that are simultaneously robust against noise and leaks very little, if any, efficiency under the standard (unperturbed) situation.  Section \ref{sec:dest} develops a new \textit{robust} learning theory to tackle this important issue.
\vskip.2em

~~$\bullet$ Fundamentals of Applied Statistics:
The new mechanics of data modeling that is presented here is not a ``one-trick pony,'' it has  deep connections with the basic fundamental statistics.  In section  \ref{sec:unif}, we demonstrate how one can systematically derive and generalize (to large high-dimensional problems in a scalable and flexible manner) many traditional and modern statistical methods as a special case of our framework. This could be especially useful for the practice and pedagogy of ML-powered modern applied statistics.  Ultimately, interlinking Statistics and Machine Learning will enrich and revitalize both the communities, by creating excitement to work across the boundary.

%: For example, in Section \ref{sec:statML} we show that how machine learning can help to uplift the conventional statistical modeling techniques from small-data . 
%\vskip.2em
%~~$\bullet$ Connection with Fundamentals of Statistics:  

%%%%%%%%%%%%%%%%%%%%%%%%%%%%%%%%%%%%%%
\section{Integrated Statistical Learning Theory} \label{sec:theory}
In this section, we introduce the basic theoretical underpinnings, principles, and methods of the integrated statistical-learning framework.
%Principle that enables

%%%%%%%%%%%%%%%%%%%%%%%%%%%%%%%%%%%%%%%%5
\subsection{Entropy, Uncertainty, and Prediction} \label{sec:uip}

The marginal distribution $f_Y(y)$ embodies uncertainty of $Y$. Now, once we observe $X=x$, the uncertainty is updated to $f_{Y|X=x}(y)$. We say that some information is received through the observation $X=x$, if the conditional distribution $f_{Y|X=x}(y)$ differs from the initial $f_Y(y)$. Therefore, one can measure the \textit{additional} information gain on $Y$ after observing $X=x$ by examining the \textit{change} in the probability distributions $f_{Y}(y)$ and $f_{Y|X=x}(y)$. Information theory provides an elegant framework for quantifying this ``change.'' 

\begin{defn} \label{def:duM} The conditional information density function (the reason behind this name will be clear soon) between $F_Y$ and $F_{Y|X=x}$ is defined as
\beq \label{eq:duM}
d(u;F_Y,F_{Y|X=x}):=\,d_x(u)\,=\dfrac{f_{Y|X=x} ( F_Y^{-1}(u))}{f_Y( F_Y^{-1}(u))},~~0<u<1\eeq
which satisfies: $\int_0^1  d(u;F_Y,F_{Y|X=x}) \dd u=1$. To improve readability, $d(u;F_Y,F_{Y|X=x})$ will be abbreviated as $d_x(u)$, with the understanding that this \textit{compares} marginal with conditional. 
\end{defn} %To make our notation less cumbersome, For the sake of brevity

\begin{defn} \label{def:kl}
Kullback-Leibler (KL) divergence between $f_Y(y)$ and $f_{Y|X=x}(y)$ is defined as
\beq \label{eq:kl}
{\rm KL}(f_{Y|x};f_Y)= \int f_{Y|x}(y) \log \frac{f_{Y|x}(y)}{f_Y(y)} \dd y.
\eeq
\end{defn}

\begin{thm} \label{thm:kl}
The KL-divergence between $f_Y(y)$ and $f_{Y|X=x}(y)$ can be rewritten as KL-divergence between $d_x(u)$ \eqref{eq:duM} and uniform density over $[0,1]$:
\beq \label{eq:kld}
{\rm KL}(f_{Y|x};f_Y)~= ~{\rm KL}(d_x;\mathbbm{1}_{[0,1]}).
\eeq
\end{thm}
{\bf Proof}: First substitute $F_Y(y)=u$ in \eqref{eq:kl}, and then apply the definition \eqref{eq:du} to deduce:
\beq \label{eq:kld2}
{\rm KL}(d_x;\mathbbm{1}_{[0,1]})~=~\int_0^1 d_x(u) \log d_x(u) \dd u.\eeq
\begin{rem}
Theorem \ref{thm:kl}  formally implies that the deviation of $d_x(u)$ from uniformity (see figure \ref{fig:bfly}(c) and visually compare the blue curve with the black dotted uniform line) captures the amount of (predictive) information gained on $Y$ by virtue of a priori knowing $X=x$. For that reason, we call $d_x(u)$  the ``conditional information density.'' Consequently, it is not surprising that $d_x(u)$ will play a fundamental role in developing our statistical theory of prediction. 
\end{rem}
For each $x_i$ in our dataset, we will get a different $d_{x_i}$ and a different KL-divergence value \eqref{eq:kld2}. Can we measure the overall predictive power of $X$ by taking average of all these different KL-divergence values? To satisfactorily answer that, we have to introduce the concept of entropy. 
\begin{defn} Entropy is the most natural and fundamental measure of uncertainty. For a continuous random variable with distribution $g$, it is defined as $H(g)=-\int g \log g.$
\end{defn}
Predictability-index can be defined as the \textit{difference} between the entropy of $Y$ and $Y|X$, which measures how much of our uncertainty about $Y$ \textit{decreased} after observing $X$. The following theorem establishes a beautiful connection between predictability-index and our conditional information density function.
\begin{thm} \label{thm:entropydiff}
The average predictability of $Y$
with respect to $X$ (i.e., reduction in entropy) can be interpreted as how different $d_x(u)$ is from uniform distribution on average:
\beq
H(Y)- H(Y|X)~ = ~\Ex_X\big[ {\rm KL}(d_X;\mathbbm{1}_{[0,1]}) \big]. 
\eeq
\end{thm}
The proof is given in Appendix A.3. The important point here is that predictive information can be expressed solely in terms of $d_x$, which distills the discrepancy (the ``gap'') between the marginal $f_Y$ and the conditional $f_{Y|x}$. Accordingly, conditional information density (CID) $d_x(u)$ acts as a \textit{natural starting point} for developing a statistical theory of predictive modeling.

%%%%%%%%%%%%%%%%%%%%%%%%%%%
\subsection{$d$-Modulated Conditional Density Representation} \label{sec:dmod}

\begin{defn}[$d$-modulation] \label{def:fd}
Any arbitrary conditional density can be expressed as 
\beq \label{eq:fd}
f_{Y|X=x}(y)\,=\,f_Y(y) \cdot d_x(F_Y(y)), \eeq
where $f_Y(y)$ is the marginal distribution of $Y$ and $d_x$ is defined in Eq. \eqref{eq:duM}.
\end{defn}

A few key points about this conditional density formula:

$\bullet$ This new representation of conditional density is universally valid and does not impose \textit{any} distributional assumptions on the outcome (e.g., dispersion, symmetry, skewness, heavy-tailedness, etc.). Most importantly, it decouples the conditional density into two components: the marginal $f_Y$ (the `pivot' density) and the conditional information density $d_x$ (which encapsulates all the essential knowledge for predicting $Y$ from $X=x$).

%Density modulation technique: 
$\bullet$ We call the density perturbation formula of Eq. \eqref{eq:fd} the $d$-modulation of marginal $Y$. The function $d_x$ \textit{filters} out the directions by which the initial $f_Y(y)$ departs most significantly from the true $f_{Y|X=x}(y)$. For that reason, we call it, alternatively, the \textit{contrast} density function. 
 
We now provide another interpretation of $d_x(u)$ by introducing the notion of conditional probability integral transform  (shortly, \texttt{cPIT}).

 \begin{defn} 
For continuous $Y\sim F_Y$ and $Y_{|x} \sim F_{Y|X=x}$, the following random variable 
 \beq U_{|x}~=~F_Y(Y_{|x})\eeq
 is defined as conditional probability integral transform. Note that when  $F_Y=F_{Y|X=x}$, i.e., under homogeneity,
$U_{|x}$ follows uniform distribution.
 \end{defn}

\begin{thm}
$d_x(u)$ is the distribution of $F_Y(Y_{|x})$, where, recall, $F_Y$ is the marginal cdf of $Y$ and $Y_{|x}$ denotes the conditional random variable $Y|X=x$.
\end{thm}
{\bf Proof}: $D_x(u)=\Pr(F_Y(Y_{|x}) \le u)=F_{Y|x}(  F_Y^{-1}(u))$. Taking derivative we get the density:
\[ D'_x(u) \,\equiv\, d_x(u)\,=\, \dfrac{f_{Y|x}(F_Y^{-1}(u)) }{f_Y(F_Y^{-1}(u))},~0<u<1.\]
\begin{rem}
Definition \ref{def:fd} uses marginal distribution as a natural pivot for representing conditional density of $Y$ given $X=x$. However, one can choose \textit{any} reasonable density function (parametric, nonparametric, etc.) as a pivot, as long its support\footnote{Support of a distribution is the region where probability density (or mass function) is strictly positive.} contains the support of $f_{Y|X=x}(y)$. The generalization of the formula \eqref{eq:fd} for arbitrary pivot $f_0(y)$ is straightforward:
\beq \label{eq:gfd}
f_{Y|X=x}(y)\,=\,f_0(y) \cdot d_x^0(F_0(y)), \eeq
where $d_x^0(u)$ is the contrast density between $F_0$ and $F_{Y|X=x}$ as defined in \eqref{eq:du}.
\end{rem}
%This is a flexible \textit{class} of models

%nucleus

%Under homogeneity, when $F_{Y|x}=F_Y$, $F_Y(Y_{|x})$ becomes an uniform random variable with density $d_x(u)=1$ for $0<u<1.$

\begin{rem}(Herbert Simon's theory to simulate human thinking) %Machine to Simulate Human Thinking
%How to simulate human thinking? 
In the year of 1956, at the historic conference ``Dartmouth Summer Research Project on Artificial Intelligence,’’ Nobel laureate Herbert Simon, along with Cliff Shaw and Allen Newell, presented a universal problem solver machine called `General Problem Solver' [GPS], which is considered by many to be the first AI program. Their theory is based on the idea of ``Difference Machine,'’ which proceeds as follows: (i) compare \texttt{C} [initial steady-state] with \texttt{F} [future desired-state] to check whether there exists any difference;
(ii) devise some technique (``intelligence'') to
 rectify the present state to reduce the important differences between \texttt{C} and \texttt{F}--call it the difference-operator; (iii) if this succeeds, then we have created a new object \texttt{A}$'$, which (one hopes) no longer has the difference when compared with the desired state \texttt{F}. The question remains, how can we transform this hierarchical logic into a computable theory? It's quite fascinating that the GPS logic is already encoded into our modeling scheme \eqref{eq:fd}, where the contrast function $d_x$ acts according to Herbert Simon’s difference-operator. In our setup, $d_x$ actively works to remove the differences between the present (steady/null) state $f_Y(y)$ and the future (desired) state $f_{Y|X=x}(y)$. The main reference here is \cite{ newell1959report}. Similar ideas also appeared in the Ph.D. thesis of Patrick \cite{winston1970thesis}, and more recently Marvin \citeauthor{minsky2007book}'s (2007) latest book on building an AI-system with commonsense reasoning. 
\vspace{-.6em}
\end{rem}
%\footnote{On common sense knowledge and reasoning, John McCarthy (2007) said ``This is the area in which AI is farthest from human-level, in spite of the fact that it has been an active research area since the 1950s''}

%%%%%%%%%%%%%%%%%%%%%%%%%%%
\subsection{A Robust Learning Theory} \label{sec:dest}
It is evident from the decomposition \eqref{eq:fd} that we need to estimate  $d_x(F_Y(y))$ to obtain the conditional density of $Y$ given $X=(X_1,\ldots,X_p)$. In this section, we describe a special technique for estimating $d_x$ that enjoys the following desirable properties:

~~$\bullet$ Robust as well as efficient: It has been already noticed that (i) existing statistical learning methods are notoriously sensitive to noise, which makes them an easy target for adversarial attacks; and at the same time, (ii) naively constructed robust algorithms perform poorly in the ideal setting with no contamination. The real question comes down to this: \textit{how} to robustify algorithms so that it can be resistant to noise/outliers while also making sure it is sufficiently accurate (efficient) when applied to relatively clean data.\footnote{This `robustness-efficiency' dilemma (also known as `rank-and-size' duality) is not new. It is at least $60$ years old unsettled issue of Statistics; see \cite{box1953non} and \cite{tukey1960}.}  Here we will present a new \textit{class} of statistical learning methods to balance robustness and efficiency. 

~~$\bullet$ Scalable to massive high-dimensional data: An important practical goal is to have a method that works for large-$(n,p)$ problems. We achieve this goal by utilizing the power of general machine learning tools. 

~~$\bullet$ Interpretable: It would be nice to have a tractable (and preferably smooth) model for the contrast density $d_x(u)$ with a few interpretable parameters that can capture increasingly complex shapes of the conditional distributions. 
%%%%%%%%%%%%%%%%%%%%%%%

Its a challenging task to reliably learn the ``true'' relationship between the input features $X$ and output $Y$ from big noisy datasets. To ensure that we succeed at it, our method has to be `doubly-robust,' and thus tackle possible contamination in \textit{both} response and covariates. We will approach the estimation of $d_x$ in two stages: first, we will make it robust with respect to the outcome variable $Y,$ and then we will discuss defence strategies to guard against corrupted input $X$.

{\bf $Y$-Robustness}. Substitute $u=F_Y(y)$ in the definition \eqref{eq:duM}, to see that $d_x(F_Y(y))$ allows us to represent the ratio $f_{Y|X=x}(y)/f_Y(y)$ as a function of the rank-transform (i.e., probability integral transform) $F_Y(Y)$. This is the first \textit{major} step towards ensuring the $Y$-robustness of our method. As a result, one can expand $d_x(F_Y(y))$ in the orthonormal basis of $F_Y(y)$. One such orthonormal system is the LP-family of rank-polynomials \citep{Deep17LPMode,deep16MetaLP,D20copula,DeepLPKsample19}, which we denote $\{T_j(y;F_Y)\}$ to emphasize that they are polynomials of $F_Y(Y)$ \textit{not} $Y$, hence extremely robust. As the true $F_Y$ is unknown, we will instead use the empirical LP-bases (eLP) $\{T_j(y;\wtF_Y)\}$ for our data analysis. Appendix A.2 describes its construction and properties. 

Summing up the above discussion, we have the following LP-series representation: 
\beq \label{eq:dLP}
d_x(F_Y(y))~=~1\,+\,\sum_{j} \LP_{j|x} T_j(y;F_Y).
\eeq
The problem of estimating the function $d_x$ now boils down to estimating the orthogonal coefficients $\LP_{j|x}$. To get an expression for the conditional LP-coefficients, first recall that $T_j$'s are mean zero random variables and are orthonormal with respect to measure $F_Y$, i.e., 
\beq \label{eq:onor}
\int T_j(y;F_Y) T_k(y;F_Y) f_Y(y) \dd y~=~ \delta_{jk},~\eeq
where $\delta$ is the Kronecker delta function.
Consequently, we multiply both sides of the equation \eqref{eq:dLP} with $T_j(y;F_Y)\cdot f_Y(y)$ and then integrate to produce the following identity: 
\beq \label{eq:lpcof1}
\LP_{j|x} ~=~\int_{-\infty}^\infty  d_x(F_Y(y))\, T_j(y;F_Y)\,f_Y(y)  \dd y, ~~\text{for $j=1,2,\ldots$}
\eeq
This immediately yields the following important theorem.
\begin{thm} \label{thm:clp}
The $j$th conditional LP-Fourier coefficient admits the following representation
\beq \label{eq:cxLP}
\LP_{j|x} ~=~\Ex\big[ T_j(Y;F_Y)|X=x   \big].
\eeq
\end{thm}
The proof is trivial; just substitute $d_x( F_Y(y))$ in equation \eqref{eq:lpcof1} by $f_{Y|X=x}(y)/f_Y(y)$. Theorem \ref{thm:clp} allows us to interpret the conditional coefficients $\LP_{j|x}$ (which are surely function of $x$) as a specialized regression function, where we regress $T_j(y;F_Y)$ on $X$. This is extremely handy, because one can now use \textit{any general} machine learning method to estimate these coefficients!

\vskip.34em
{\bf $X$-Robustness}. Nevertheless, the computational theory \eqref{eq:cxLP} is still not fully satisfactory, as corrupted (or adversarially perturbed) $X$ can substantially influence the coefficients. To provide further defense, we need one last piece of idea, which is a little-known but fundamental fact about `regression \textit{via} rank-transform.' Also see Appendix A.4.2 for something surprising.
\begin{thm} \label{thm:rankreg}
For any general random variable $X$ (whether discrete or continuous) and a measurable function $\Psi(\cdot)$, the following result holds
\beq \Ex[\Psi(Y) \mid X] \,=\, \Ex[\Psi(Y) \mid  F_X(X)],~ \text{with probability $1$}
\eeq
\end{thm}
{\bf Proof}: We start with a basic quantile mechanics fact:  for any random variable $X$ we have
\[Q_X(F_X(X))~=~X, ~~\text{with probability 1}.\]
where $Q_X(u)=\inf_x\{F_X(x) \ge u\}$, $0<u<1$ denotes the quantile function of $X$. Accordingly, any function of $X$, say $h(X)$, can be equivalently rewritten as a function of the rank-transform $F_X(X)$, since $h(X)$ is equal to $h\circ Q_X(F_X(X))$ with probability $1$.

In plain language, this result proves the ``sufficiency'' of rank-transform (retaining \textit{all} the information) for estimating conditional mean function. Therefore, we can estimate $\LP_{j|x}$ \eqref{eq:cxLP} by regressing $T_j(Y;\wtF_Y)$ on the LP-bases of $X$. This will allow doubly-robust estimation of $d_x$ \textit{without sacrificing} an iota of efficiency in the process. Next, we will go over some examples, to demonstrate how this theory actually translates into practice (algorithm). 
\vskip.35em
%inbuilt defence mechanism

\begin{figure}[t]
\includegraphics[width=.45\linewidth,trim=1cm 1cm 1cm 1cm]{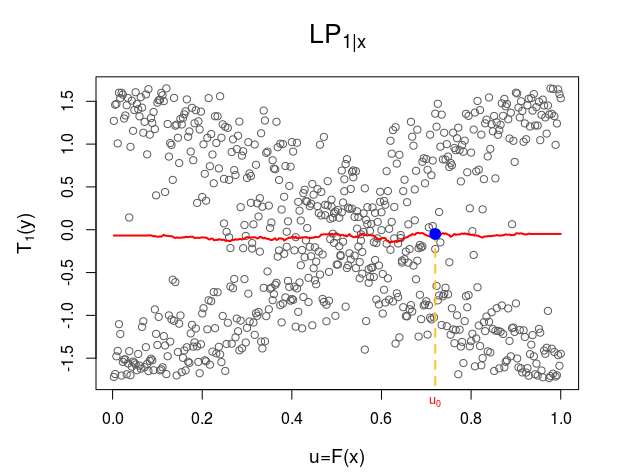}~~~~~~~~\includegraphics[width=.45\linewidth,trim=1cm 1cm 1cm 1cm]{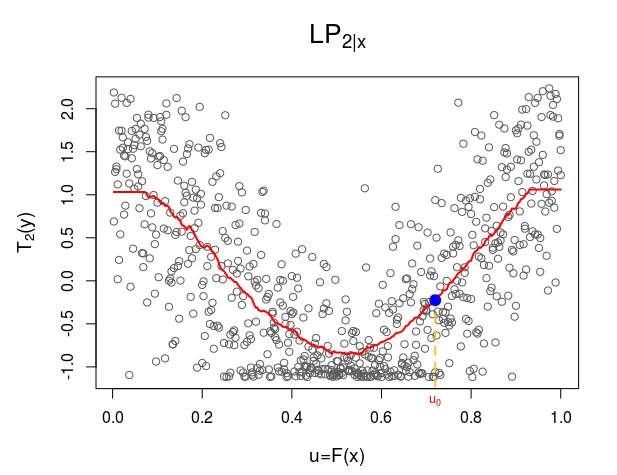}\\[3em]
\includegraphics[width=.45\linewidth,trim=1cm 1cm 1cm 1cm]{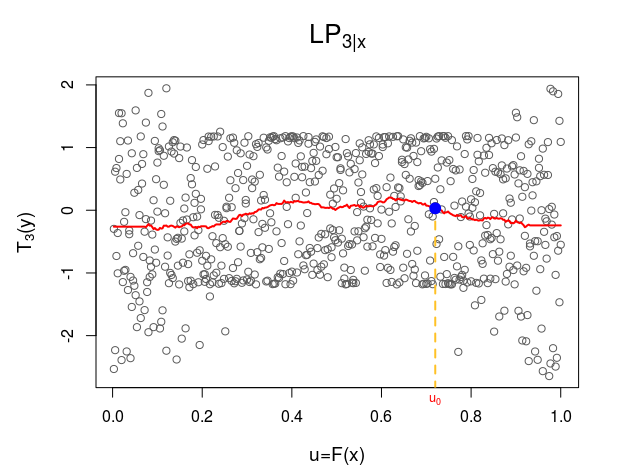}~~~~~~~~\includegraphics[width=.45\linewidth,trim=1cm 1cm 1cm 1cm]{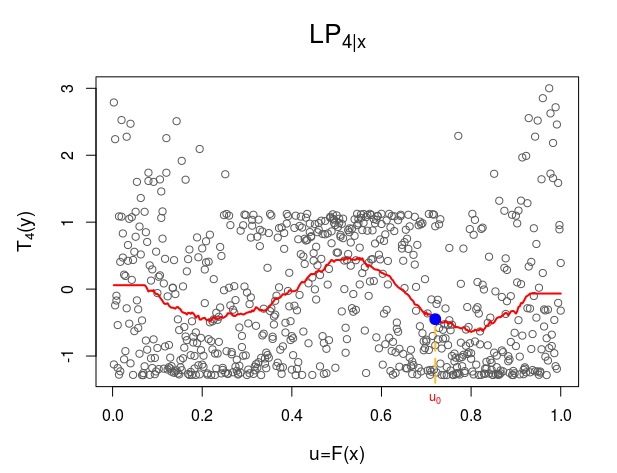}
\vskip2em
\caption{First $4$ LP-scatter plots for the butterfly data. It illustrates the steps for obtaining $\widehat\LP_{j|x}$ using generic machine learning-program, by operating on the LP-transformed domain.} \label{fig:bfly2}
\vspace{-.3em}
  \end{figure}

%%%%%%%%%%%%%%%%%%%%%%%
%\subsubsection{Illustrating with Data}
{\bf  Example 1.} \textit{The Butterfly data}. Fig. \ref{fig:bfly2} displays the scatter plots of $T_j(y,\wtF_Y)$ against $u=\wtF_X(x)$ for $j=1,\ldots,4$. We refer these graphs as LP-scatter plots. The red curves denote knn-estimated regression curves. In the \texttt{R} computing language, this is simply \texttt{knn}($T_j(y;\wtF_Y)$\, $\sim$ $\,T_X$), $(j=1,\ldots,4)$ where $T_k(x;\wtF_X)$ is the $k$th col of LP-transformed feature matrix $T_X$. Of course, instead of knn method, one can choose any machine learning routine; the next example uses gradient boosting algorithm (gbm). To get the estimated conditional LP-coefficients at $x_0=2$ (indicated by the blue dots) we record the predicted values at $u_0=\wtF_X(x_0)$: $\widehat\LP_{2|x_0}$ equals to $-.020$, $\widehat\LP_{4|x_0}$ equals to $-0.47$, and the rest are practically zero. Substituting them into \eqref{eq:dLP} yields the estimated contrast function $\whd_{x_0}$, as was shown in Fig. \ref{fig:bfly}(c).
%%%%%%%%%%%%%%%%%%%%%%%%%%%%%%
  
  %%%%%%%%%%%%%%%%%%%%%%%%%%%%%%

\begin{figure}[t]
\vskip.4em
\begin{subfigure}[t]{.345\linewidth}
    \centering
    \caption{The Data}
        \includegraphics[width=\linewidth,trim=1cm 0cm 0cm .5cm]{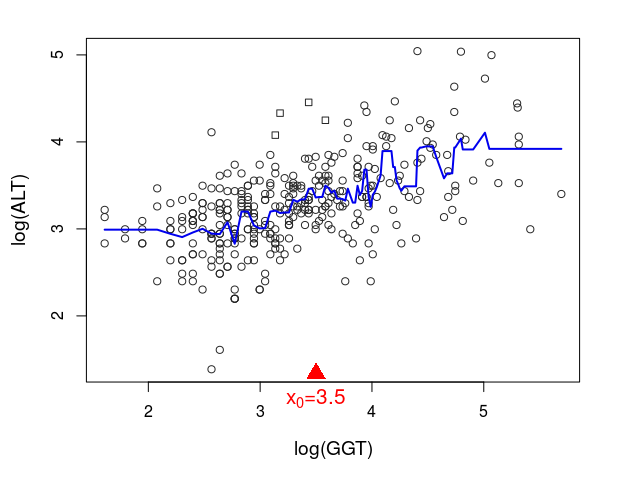}
    \end{subfigure}~
    \begin{subfigure}[t]{.325\linewidth}
    \centering
    \caption{Contrast Density}
        \includegraphics[width=\linewidth,trim=1cm 0cm 0cm .5cm]{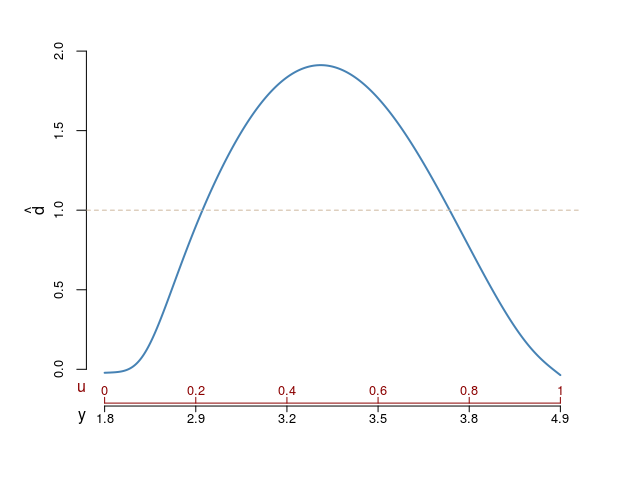}
    \end{subfigure}~
 \begin{subfigure}[t]{.340\linewidth}
    \centering
    \caption{Conditional Density}
        \includegraphics[width=\linewidth,trim=1cm 0cm 0cm .5cm]{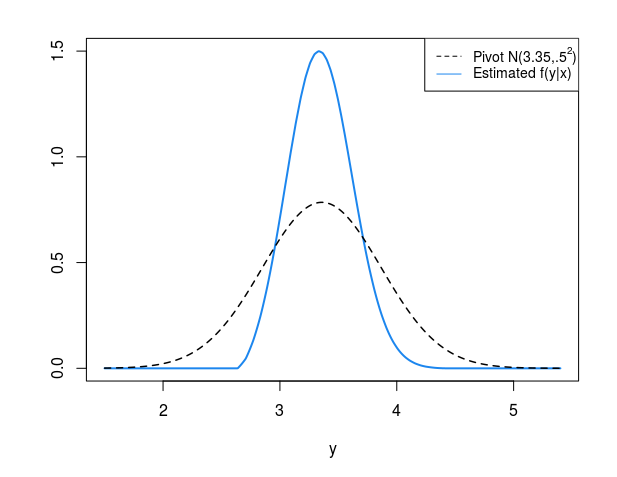}
    \end{subfigure}
\caption{BUPA liver data: (a) The ``wiggly'' blue line denotes the fitted gbm regression curve; (b) the estimated contrast density function $d_{x_0}\equiv d(u;F_0,F_{Y|X=x_0})$. It is asymmetric and inverted ``U''-shaped; (c) the estimated $\hf_{Y|x_0}(y)$ along with the starting pivot $f_0 = \cN(3.35,.5^2)$.}  \label{fig:bupa}
\vspace{-.4em}
  \end{figure}

{\bf  Example 2.} \textit{BUPA liver disorders data}. This is a popular dataset (available in the UCI ML repository) used by many researchers, including \cite{breiman01}. It consists of measurements of gamma-glutamyl transpeptidase (GGT) and alanine-aminotransferase (ALT) extracted from  $N=345$ male individuals' blood samples. High levels of ALT and GGT indicate a damaged liver. Fig. \ref{fig:bupa} (a) shows the scatter plot of  $y$=log(ALT) versus $x$=log(GGT). We seek to predict the distribution of log-ALT levels given $x=3.5$; see the red triangle in panel (a) of Fig. \ref{fig:bupa}.

%%%%%%%%%%%%%%%%%%%%%%%%%%%%%%
%%%%%%%%%%%%%%%%%%%%%%%%%%%%%%
Our data analysis scheme can be simply summarized as follows:

~1) We start by modeling the conditional mean function using gradient-boosting algorithm. The rugged blue line denotes the estimated regression curve $\widehat{\mu}(x)$. 

~2) To estimate the conditional density at $x_0=3.5$, we choose $\cN(\widehat{\mu}(x_0), \si_y^2)$ as our pivot model, where $\widehat{\mu}(x_0)=3.35$ and $\si_y=0.5$.

~3) Following theorems 4 and 5, we estimate the conditional LP-Fourier coefficients $\LP_{j|x_0}$ and the associated contrast density function: $\whd_{x_0}(F_0(y))~ \approx ~1 - 0.72\, T_2(y;F_0) + 0.10\, T_3(y;F_0).$
This is shown in the middle panel of Fig. \ref{fig:bupa}.

~4) Interestingly, we can interpret the coefficients of $\whd_{x_0}$: the large value of $\widehat \LP_{2|x_0}$ indicates that the selected pivot needs scale-correction (``second-order'' correction) to get close to the true conditional density at $x_0$. In addition, the negative sign implies the variability has to be reduced, which is vividly apparent from Fig. \ref{fig:bupa}(c).

~5) Similarly, the presence of third-order (positive) LP-coefficient indicates the symmetric normal pivot needs to be tilted to the right. The positive skewness of $f_{Y|x_0}(y)$ is probably justified due to the presence of few large $y$-values\footnote{Classical robust methods treat them as ``outliers'' and devise mechanisms for removing them from analysis. This is incorrect. A careful examination will reveal that these large-$y$ values (indicated by squares) have some pattern; they are not randomly scattered (like outliers). The real question of what is an outlier and what is a ``real'' pattern, finally boils down to artfully balancing robustness with efficiency.} around $x_0=3.5$, as indicated by squares.

~6) Finally, the shape of $\whd_x$ provides quick exploratory guidance for choosing appropriate families of parametric conditional distributions. For the BUPA liver-disorders data, one could select skew-normal or asymmetric logistic distribution \citep{FriedmanCD2020} as a model for $Y_{|x}$.

%%%%%%%%%%%%%%%%%%%%%%%
\subsection{The Basic Algorithm} \label{sec:gALGO}
We are now in a position to provide the blueprint of the algorithmic interface in Fig. \ref{fig:upm_flowchart} that converts a user-selected machine learning method into an uncertainty prediction machine in a robust and interpretable manner. The theory in the previous section, makes it absolutely straightforward and extremely easy to implement, which essentially involves three steps. The following algorithm presents a flowchart of the core computational steps in pseudo-code, so that one can implement it using \textit{any} programming language or ML-platform. We have found that the H2O open-source is extremely efficient. But users can choose any ML environment (e.g., TensorFlow, Azure, AutoGloun, etc.) to operationalize the following workflow.
\begin{center}
\vspace{-.3em}
\hskip1.5em Interface Design: A flowchart of the core computational steps in pseudo-code
\end{center}
\vspace{-.8em}
\medskip\hrule height .65pt
\vskip.25em
~~{\bf Input} Data $(X_i,y_i)$ for $i=1,\ldots, N$; user-selected machine learning method (\texttt{ML}); values of $m_y$ and $m_x$. And the target test-point $x_0 \in \cR^p$. 

~~Step 1. $T_Y \leftarrow$ \texttt{LP.basis}$(y,m_y)$; $T_X \leftarrow$ \texttt{LP.basis}$(X,m_x)$

~~\texttt{for}($j=1,\ldots,m_y$)

~~\{

~~~~Step 2. \texttt{Fitted.ML}$_j$ $\leftarrow$~ \texttt{ML}\big($T_Y[\,, \,j] \sim T_X$\big)

~~~~Step 3. $ \LP_{j|x_0} \leftarrow$~ \texttt{predict}(\texttt{Fitted.ML}$_j$, $x_0$)~

~~\}

~~{\bf Return} $\whd_{x_0} \,\leftarrow \,1 + \sum_j \LP_{j|x_0}  T_Y[\,, \,j]$

\medskip\hrule height .65pt
\vskip1em

The LP-transformation is implemented by the function \texttt{LP.basis}; $T_X$ is the stacked matrix $[T_{X_1}\mid T_{X_2} \mid \cdots \mid T_{X_p}]$, where the $j$th column of the matrix $T_{X_\ell}$ is simply $T_j(x;\wtF_\ell)$. Step 1 makes the whole procedure robust, non-linear, and \textit{invariant} under monotone transformations. Step 2 performs the training of the ML method  and step 3 predicts the fitted value at the desired $X=x_0$. Based on our experience, $m_x=4$ and $m_y=6$ work reasonably well for most problems, including all the datasets analyzed in this paper. Finally, one can generate the uncertainty distribution $f_{Y|x_0}$ by applying the $d$-modulation formula of equation \eqref{eq:fd}.

%%%%%%%%%%%%%%%%%%%%%%%%%%%%%%%%%%%%

\begin{rem}
In this section we have showed: how to actualize the vision sketched in Fig. \ref{fig:upm_flowchart};
how to compress-down large complicated algorithmic methods into concise parametric forms with few interpretable coefficients\footnote{This materialize the objective of `Trend 1'' as proposed by  \cite{efron2020pred}: ``Trend 1 aims to make the output of a prediction algorithm more interpretable, that is, more like the output of traditional statistical methods.''}; how to derive a smoothness-promoting model for the response $Y$ from black-box prediction methods\footnote{This operationalizes what \cite{efron2020pred} envisioned: ``Smoothness of response is not built into the pure prediction algorithms...The natural desire to use them for scientific investigation may hasten development of smoother, more physically plausible algorithms.''}; and how to robustify algorithmic techniques\footnote{\cite{FriedmanCD2020} advocates strongly for this.}. We have achieved each of these ends by devising a generic mechanism that holds for \textit{any} machine learning method. Furthermore, Section \ref{sec:unif} shows how these same modeling principles can unify a wide range of traditional and novel statistical methods.
\vspace{-.35em}
\end{rem}
% Generic applicability of these principles.

% The only contribution here is to: able to extract a simple interpretable (parametric) \textit{smooth} and robust model out of ..

% How can we build a statistical model (hopefully robust and parametric for easy interpretation) starting with a black-box algorithmic one?

% {\bf Algorithmic Model Condensation}: 
% here we will describe how a complex ML method can be converted into c. Nonparametric principle. 
% It compress-down a large-scale ML algorithm into few interpretable statistical parameters

% ~~$\bullet$ 

%~~$\bullet$ expressive power: Any higher-order shape analysis for more complete understanding. At the core the learned $d_{Y|X=x}$-function plays an important role. 

%parametric modeling culture: We construct a parameterizable model `looking at the data' instead of assuming...that can explain the data generating process

%two “approaches” of parametric and algorithmic modeling.
%A central need in all of statistics and machine learning is then to develop the tools and theories that allow 

%{\bf Example} https://arxiv.org/pdf/1707.03307.pdf

%Section 6.2: location ($x+x^2$) with scale being $1+sin(2x)$ and also vary the skewness/shape parameter $x$ (from $-4$ and $4$); x is uniform$[-4,4]$. We simulate $n = 1000$ data points.. 

%https://en.wikipedia.org/wiki/Skew_normal_distribution

%%%%%%%%%%%%%%%%
\subsection{$d$-Kernel Smoothing and Importance Weighting} \label{sec:iw}
In many applications, we are often interested in estimating quantities like  $\Ex[\Psi(Y)|X=x]$ for general $\Psi(\cdot)$ function. This can be rewritten as
\[\Ex[\Psi(Y)|X=x]\,=\int_y \Psi(y) f_{Y|X=x}(y)\dd y\,=\int_y \Psi(y) f_Y(y) \,d_x(F_Y(y)) \dd y,\]
by applying the formula \eqref{eq:fd}. This leads to the following theorem, which expresses the (local) conditional mean as a weighted average of the global samples $\{y_1,\ldots,y_N\}$.
%%%%%%%%%%%%%%
\begin{thm}[$d$-kernel smoothing] \label{thm:dker} Conditional expectation of $\Psi(Y)$ given $X=x$ can alternatively be expressed as:
\beq \label{eq:cex}
\Ex[\Psi(Y)|X=x]=\Ex\big[\Psi(Y) \cdot d_x\big(F_Y(Y)\big)\big],
\eeq
the rhs expectation is taken over the marginal distribution of $Y$. 
\end{thm}
\begin{rem}
The above representation implies the following $d$-weighted estimator:  
\beq \label{eq:cexest}
\widehat{\Ex}[\Psi(Y)|X=x]\,=\,\dfrac{1}{N}\sum_{i=1}^N w_i(x) \Psi(y_i),
\eeq
where the empirical weighting function $w_i(x)=\widehat{d}_x( \wtF_Y(y_i))$. The formula \eqref{eq:cexest} can be interpreted as a specially-designed kernel smoothing, which could be extremely handy in some cases. Three particular examples are given below: 

~1)  $\Psi(Y)=\ind(Y>k)$ gives the conditional probability $\Pr(Y>k|X=x)$; This can be used to construct different risk measures based on tail events; 

~2)  $\Psi(Y)=Y$ recovers the conditional mean $\Ex[Y|X=x]$; 

~3) $\Psi(Y)=(Y-\Ex(Y))^2$ yields conditional variance $\Var(Y|X=x)$, whose square-root, i.e., the standard error of $Y_{|x}$, is often used as an uncertainty quantification measure for a prediction. However, we should point out that standard error is not always a prudent choice for quantifying reliability of a prediction, specially when the conditional distribution is  heavy-tailed, or skewed, or multi-modal. Length of prediction interval (see Section \ref{sec:PI}) provides a more robust measure. To drive this point home, example 12 discusses an interesting real prediction problem (the ``Auto MPG'' data).
\vspace{-.45em}
\end{rem}
%%%%%%%%%%%%%%%%%%%%%%%%%%%%%%%%%%%%%%%%%%%%%

%%%%%%%%%%%%%%%%%%%%%%%%%%%%%%
\section{Exploratory  Learning Machine} \label{sec:xml}
Our goal here is to develop some graphical diagnostic methods to make a transition from a \textit{black-box} answer machine to an \textit{exploratory} learning machine that can drill deeper and generate previously unanticipated, interesting questions to facilitate data-driven scientific discovery and understanding.\footnote{A `better-question' is often \textit{better} than a `better-answer.'} This can be viewed as a pre-deployment ``health check-up'' for statistical learning methods that comes with a set of nonparametric exploratory tests.

% \begin{itemize}
%     \item ML-uncertainty: goal is unmodeled ``excess'' heterogeneity identification that is invisible to the deployed ML algorithm.
% \item Pivot-uncertainty: In which way the starting pivot $f_0(y)$ in incomplete? What are the deficiency? 
% \item UPM-uncertainty: Whether the final distribution prediction machine fits the data well. to gauge the fitness level of the algorithm. 
%\end{itemize}
%%%%%%%%%%%%%%%%%%%%%%%%5
\subsection{Heterogeneity Component Analysis} \label{sec:homog}
Heterogeneity component analysis (HCA) is a technique for identifying \textit{dynamic} components of $Y_{|x}$--the aspects of the conditional distribution $f_{Y|X=x}(y)$ (such as location, scale, skewness, tail, etc.) that vary with covariates. We illustrate the main idea using the \texttt{butterfly} example. The procedure starts with LP-scatter plots, as was shown in Fig. \ref{fig:bfly2}. The first scatter plot captures how the location of $f_{Y|X=x}(y)$ (conditional mean) changes with $x$; the second one captures the change of scale (the nature of heteroscedasticity); the third one provides information on the changing skewness and so on. 
Thus, an obvious approach for detecting the heterogeneity components would be to:

~Step 1. Compute a goodness-of-fit statistic for the $j$th LP-scatter plot by its  coefficient of multiple correlation:
\[R^2_j\, =\, \text{Proportion of variance in $T_j(y;F_Y)$ that is explained by $\{T_1(x;F_X), \ldots, T_m(x;F_X)\}$}\]
~Step 2. Assess the significance of the $R^2_j$ by constructing F-statistic for the $j$th LP-scatter
\[ F_j = \dfrac{R^2_j}{1-R^2_j}\times\dfrac{N-m-1}{m}\]
whose null distribution\footnote{To allow non-normal errors, use the fact that $m F_{m,N-m-1}$ approaches to $\chi^2_m$ (in the sense of convergence in distribution) as $N \rightarrow \infty$ and compute the p-value based on this large-sample chi-square approximation.} is $F_{m,N-m-1}$. Fig. \ref{fig:bfly}(a) displays the significant (whose pvalue $<0.05$) heterogeneity components for the butterfly data. 
%--a basic common sense approach
\vskip.25em
{\bf The ``Flatland'' Hypothesis}. For real data analysis, one can apply the above method 
on residuals to determine which shape parameters are varying with $x$, beyond first-order mean. The ``flatness'' of residuals implies the errors are homogeneous with respect to the covariates.

\begin{figure}[t]
    \centering
    \begin{subfigure}[t]{.45\linewidth}
    \centering
    \caption{Bone mineral density data}
        \includegraphics[width=\linewidth,trim=1cm 0cm 0cm .5cm]{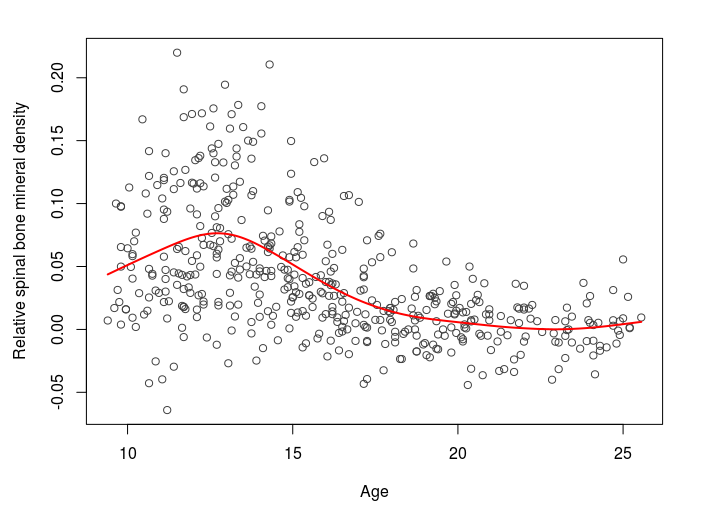}
    \end{subfigure}~~~~~~~
    \begin{subfigure}[t]{.45\linewidth}
    \centering
    \caption{HCA plot: mineral density data}
        \includegraphics[width=\linewidth,trim=1cm 0cm 0cm .5cm]{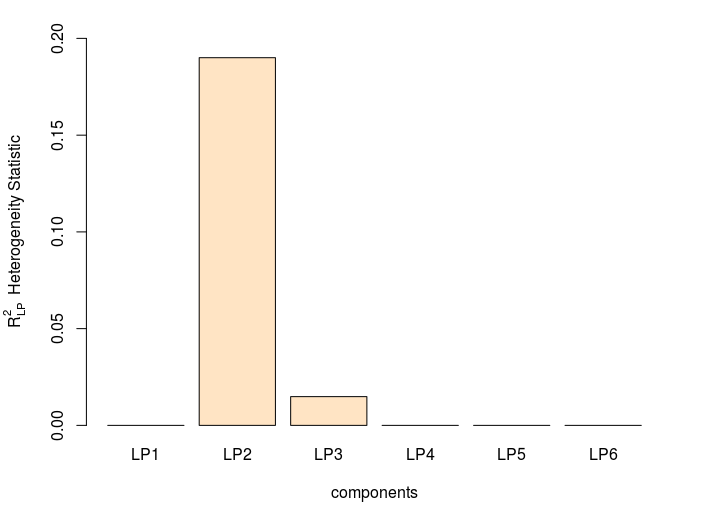}
    \end{subfigure}\\[1em]
    \begin{subfigure}[t]{.45\linewidth}
    \caption{Standardized residuals: online news data}
       \includegraphics[width=\linewidth,trim=1cm 0cm 0cm .5cm]{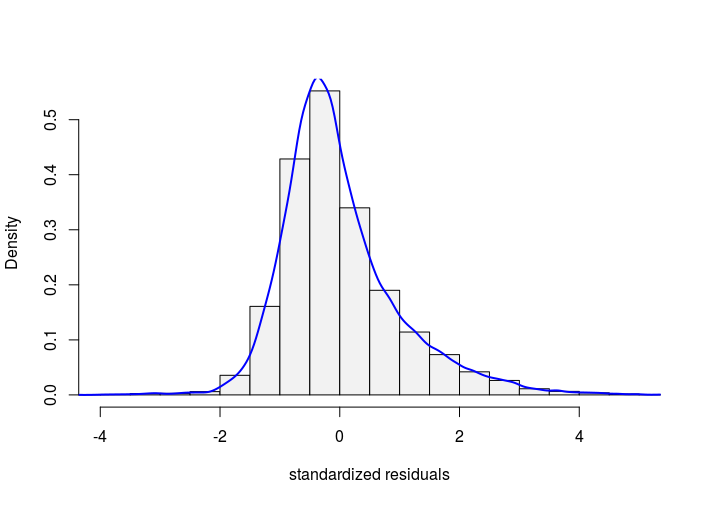}
    \end{subfigure}~~~~~~~
    \begin{subfigure}[t]{.45\linewidth}
    \centering
    \caption{HCA plot: online news data}
     \includegraphics[width=\linewidth,trim=1cm 0cm 0cm .5cm]{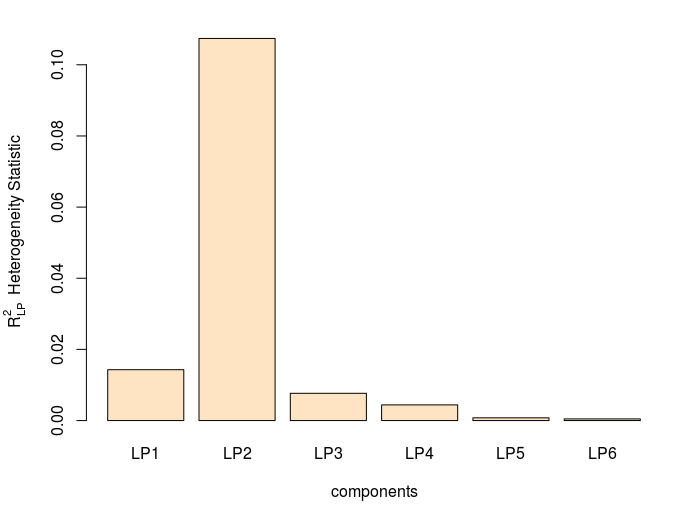}
    \end{subfigure}
    \vskip.65em
    \caption{Heterogeneity component analysis (HCA): Top row shows the bone mineral density data. Beyond conditional mean (regression), the scale and skewness (symmetry) of the error distribution are also changing with $x$. Bottom row is the online news popularity data. Left plot shows the distribution of the standardized residuals and the right one is the associated \texttt{HCA} plot. The variability needs to be modelled as a function of the covariates for better fitting.}
    \label{fig:hcomps}
\end{figure}

{\bf Example 3.} \textit{Bone mineral density data} \citep{hastie1999bone}.  It contains measurements on the relative change in spinal bone density (over one year period) of $N=485$ North American adolescents, as a function of age. This dataset was previously analyzed by \citet[p. 152]{stanfybook}. Fig. \ref{fig:hcomps}(a) shows the data along with the estimated regression smoother.  We like to understand whether there is any ``excess'' heterogeneity beyond the 1st-order regression modeling. To check the homogeneity of the residuals with respect to the covariate $x$, we apply our \texttt{HCA} procedure on the residuals $y_i-\widehat y_i$. The result is shown in the top row of Fig. \ref{fig:hcomps}, which says that the regression model has not taken into account the heteroscedasticity of the data (furthermore, there is some evidence of varying skewness in the conditional distribution). These unmodelled (dynamic) heterogeneity components should be modeled to improve the prediction. The heterogeneity is partly due to the presence of both male and female youngsters' in the dataset. 
\vskip.15em

\vskip.35em
{\bf Example 4.} \textit{Online news popularity data} \citep{fernandes2015}. The objective of this study is to predict the popularity of online news. The data\footnote{The article in row \# $31,038$ seems to be an outlier (in fact it is an erroneous data: check its 4th and 5th feature values--rate of unique and non-stop words--which can't be more than 1, by definition). Since our method is \textit{doubly-robust} we don't worry that much. But standard machine learning methods might get unduly influenced by this one data point (article). The practice of looking at the data is always a good habit.} consist of $N = 39,644$ articles, and for each article we have $y$ (log$_{10}$ of number of shares in social networks, a measure of popularity) and $p=59$ extracted features (such as number of words in the title, number of links, number of images, etc.) $x$. This dataset was recently analyzed by \cite{FriedmanCD2020}.

We start by fitting gradient boosting regression to the data. At this point, it is a natural question to ask whether the gbm captured most of the important information in the data. If not, what is missing? To spot the  unmodelled dynamic components, we start from the residuals, as shown in Fig. \ref{fig:hcomps}(c). The \texttt{HCA} diagnostic plot of panel (d) is constructed based on the  Lasso-selected features, which clearly indicates the presence of a strong heteroscedasticity. However, it is also true that the conditional distribution will be asymmetric and long-tailed, but our analysis shows that they are not swiftly changing with covariates; they are more or less static in nature, not dynamically evolving shape-parameters. This further corroborates the findings of \citet[Sec. 6.2]{FriedmanCD2020}.

%How to rectify them?

%How accurately the model describes the data generating process? If not, what was missed? 

%%%%%%%%%%%%%%%%%%%%%%%%%5
\subsection{Pivot Uncertainty Modeling}
The parameters $\LP_{j|x}$ of the contrast density $d_x$ can be viewed as ``generalized coordinates'' that describe the shape of $f_{Y|X=x}(y)$ relative to some reference distribution (pivot) $f_0(y)$. 

\newpage
%%%%%%%%%%%%%%%%%%%%%%%%%%%%%%%%%%%%%%%%
\begin{figure}[!h]
    \centering
    \begin{subfigure}[t]{.48\linewidth}
    \centering
    \caption{Data}
        \includegraphics[width=\linewidth,trim=1cm 0cm 0cm .5cm]{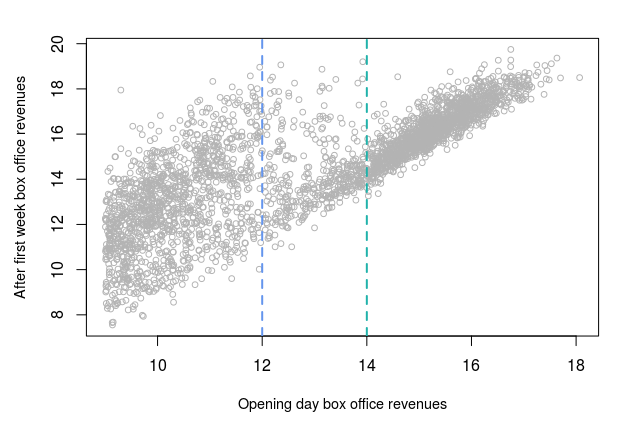}
    \end{subfigure}~~~~~~~
    \begin{subfigure}[t]{.48\linewidth}
    \centering
    \caption{Pivot}
        \includegraphics[width=\linewidth,trim=1cm 0cm 0cm .5cm]{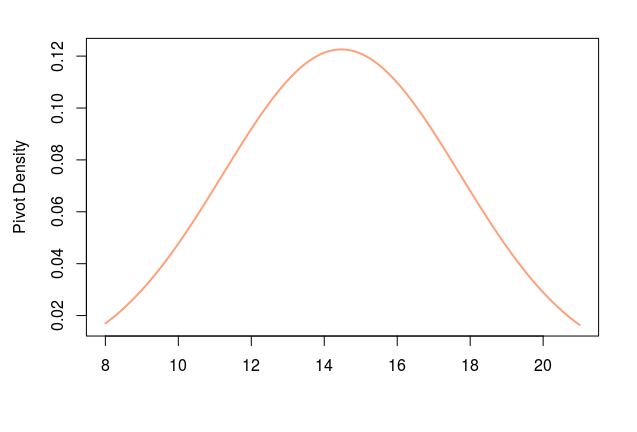}
    \end{subfigure}\\[.64em]
    \begin{subfigure}[t]{.48\linewidth}
    \centering
    \caption{Contrast Density at $x=12$}
        \includegraphics[width=\linewidth,trim=1cm 0cm 0cm .5cm]{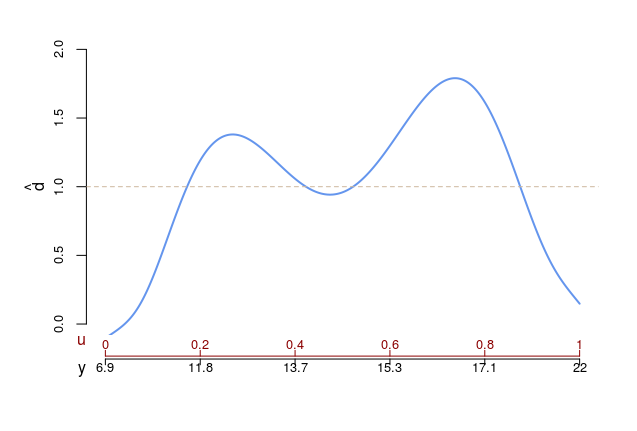}
    \end{subfigure}~~~~~~~
    \begin{subfigure}[t]{.48\linewidth}
    \centering
    \caption{Contrast Density at $x=14$}
        \includegraphics[width=\linewidth,trim=1cm 0cm 0cm .5cm]{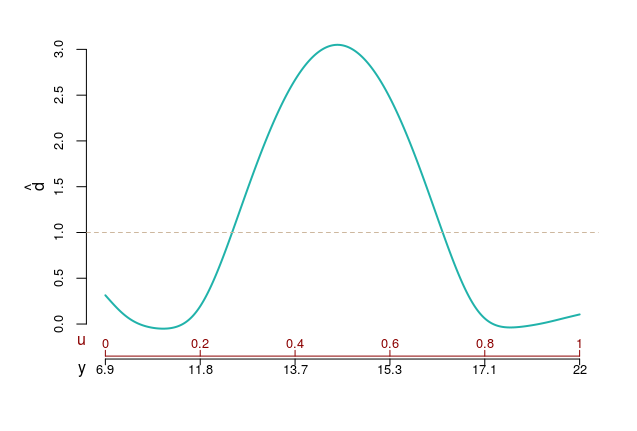}
         \end{subfigure}\\[.64em]
     \begin{subfigure}[t]{.48\linewidth}
    \centering
    \caption{Conditional density at $x=12$}
        \includegraphics[width=\linewidth,trim=1cm 0cm 0cm .5cm]{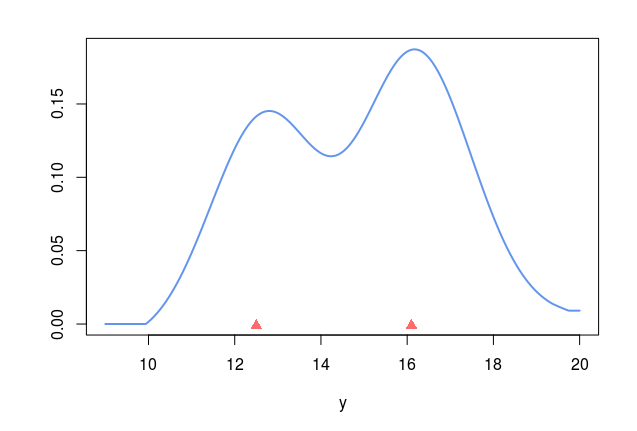}
    \end{subfigure}~~~~~~~
    \begin{subfigure}[t]{.48\linewidth}
    \centering
    \caption{Conditional density at $x=14$}
        \includegraphics[width=\linewidth,trim=1cm 0cm 0cm .5cm]{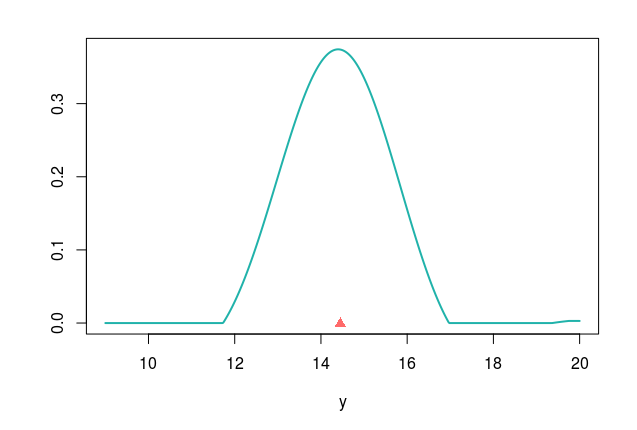}
    \end{subfigure}
    \caption{(color) $1990$'s film data: first row shows a part of the full \texttt{film90} data and the starting normal pivot. Notice the ``two-branches'' of the data cloud. Second row: The estimated contrast densities $\whd_{Y|X=x}(u)$, which describe the conversion factor: how the pivot $f_0(y)$ has to be transformed (perturb) to get to $f_{Y|X=x}(y)$. Third row: The refurbished  conditional densities $\hf_{Y|X=x}(y)$ obeying the $d$-modulation law \eqref{eq:gfd}; the red triangles denote the mode of the distributions. Notice the difference in shape between the starting Gaussian pivot (panel b) and the final estimated conditional densities (of the bottom row).} \label{fig:film}
\end{figure}
%%%%%%%%%%%%%%%%%%%%%%%%%%%%%%%%%%%%%%%
\newpage
In fact, these shape-coefficients $\LP_{j|x}$ capture the whole dynamics of how the conditional distribution $f_{Y|X=x}(y)$ evolves with $x$.\footnote{Indeed, one can even construct an $m$-dimensional phase-space (analogous to statistical mechanics), consisting of $N$ points $\{\LP_{1|x_i},\ldots,\LP_{m|x_i}\}_{i=1}^N$ to graphically represent the evolving shape of $f_{Y|X=x_i}(y)$. Each point in LP-phase-space corresponds to a particular shape for a particular value of $x=(x_{1},\ldots,x_{p})$.} Naturally, a large value of 
\beq\label{eq:qpivot}
    \qPivot(x)\,=\, \sum_{j}\big| \LP_{j|x} \big|^2
\eeq
indicates higher uncertainty (misfit) of the pivot $f_0$ for modeling the conditional $f_{Y|X=x}$. The `q' in  $\qPivot$ stands for quantification of lack-of-fit. The above formula can also be interpreted as measuring the deviation of $d_x(u)$ from uniformity (follows from Parseval's identity):
\beq\int_0^1 (d_x(u)-1)^2 \dd u~=~\sum_{j} \big| \LP_{j|x} \big|^2~~\eeq

{\bf Example 5.} \textit{Movie box-office revenue data} \citep{filmdata2012}. The goal of this study is to build a forecasting model for film revenues. The data contain $N=4031$ pairs of observations $(x_i,y_i)$ on the 1990s film data, where $x_i$ is the log of opening box-office revenues and $y_i$ is the log of box office  revenues after the first week. We are interested in predicting the distribution of the outcome $y$ for $x=12$ and $14$--marked with the blue and green dotted line in Fig. \ref{fig:film} (a). Our starting pivot $f_0 =\cN(14,\si_y=3.25)$ is shown in the panel (b). To check whether the conditional density at $x$ follows this presumed parametric law, we estimate the respective contrast densities; see the middle panel of Fig. \ref{fig:film}. The bimodality of $\whd_{x=12}$ indicates that the Gaussian pivot failed to capture the presence of films with two kinds of box-office earnings. This is also reflected in the pattern of the scatter plot, which is divided into \textit{two} branches. On the other hand, the inverted `U' shape (quadratic) of $\whd_{x=14}$ implies that the variability of the Gaussian pivot needs to be adjusted ($2$nd-order correction) to go from $f_0(y) \mapsto f_{Y|X=14}(y)$. The final refurbished  conditional density models at $x=12$ and $14$ are shown in the bottom row of Fig. \ref{fig:film}. They are estimated using the $d$-modulation technique, described in Sec. \ref{sec:dmod}.

%%%%%%%%%%%%%%%%%%%%%%%%%%%%%%%%%%%
\subsection{Goodness-of-fit Diagnostics} \label{sec:gof}
The flexibility of our approach (as depicted in Fig. \ref{fig:upm_flowchart}) enables us to choose \textit{any} machine learning method to design the uncertainty prediction machine (\texttt{UPM}). Nonetheless, a user may want to know which ML-powered distribution prediction algorithm(s) better explains the pattern in the data. To answer that here we introduce a graphical exploratory procedure to evaluate the overall goodness-of-fit of the method. Deploying statistical models without knowing how well it actually fits the data, is a statistical sin. We illustrate our idea using the \texttt{butterfly} data.
\vskip.25em

{\bf ML-driven UPM system}. To construct an \texttt{UPM} for the  butterfly data, we start with three possible base learners: k-nearest neighbors (knn), random forest (RF), and gradient-boosting algorithm (GBM). We train each ML-driven UPM model using the generic procedure described in section \ref{sec:gALGO}. To  check whether these models are congruent with the observed data, we proceed as follows: 
\begin{figure}[t]
\includegraphics[width=\linewidth,trim=1cm 1cm 1cm 1cm]{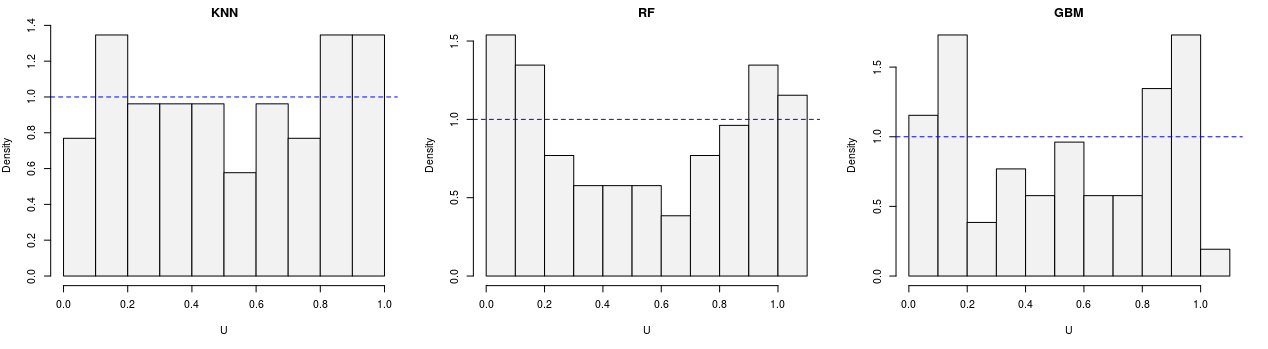}\\[3em]
\includegraphics[width=\linewidth,trim=1cm 1cm 1cm 1cm]{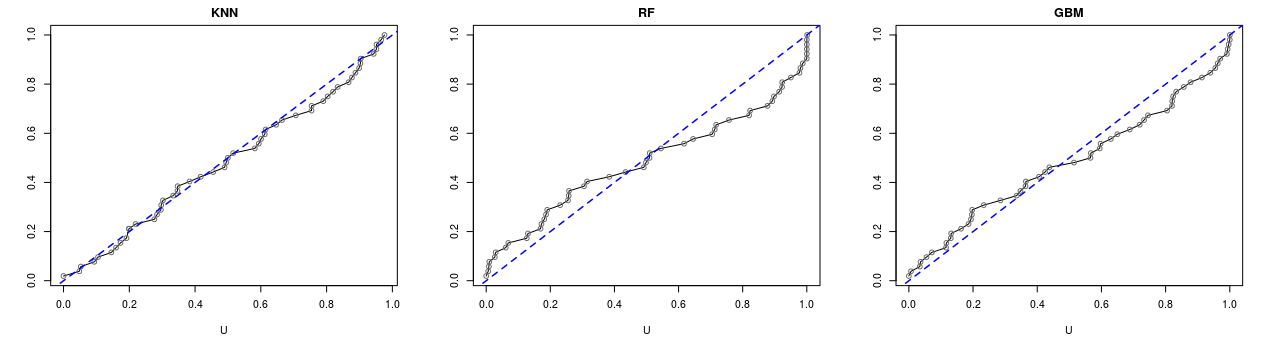}
\vskip.8em
\caption{Exploratory goodness-of-fit assessment for the \texttt{butterfly} data. The histograms and QQ plots of the generalized quantile-residuals  \eqref{eq:gqr} are shown for three ML-models. They are computed based on a 15\% hold-out data. Knn (with $k=15$) seems fits the data well.} \label{fig:gofb}
\end{figure}

~Step 1. Compute the generalized quantile-residuals for all three models on a hold-out validation dataset (separate from the training samples) of size $n_v$
\beq \label{eq:gqr} 
U_{\ell i}~=~\whF_{Y|X=x_i}^\ell(y_i),~~\text{for~$i=1,\ldots,n_v$; \,and $\ell=1,2,3$}.
\eeq
Note that if the model accurately captures the pattern in the validation data, then we expect the distribution of the $U_{\ell}$'s to be close to uniform.
\vskip.2em

~Step 2. Display the histograms and QQ-plot of $\{U_{\ell 1}, \ldots, U_{\ell n_v}\}$. This is done in Fig. \ref{fig:gofb} for the \texttt{butterfly} data, which shows the \texttt{knn} model satisfactorily captures the observed reality. 

\vskip.2em

~Step 3. To quantify the model-fit we propose the following test statistic for the $\ell$-th model: 

\beq \qDIV_\ell ~= ~\sum_{j=1}^m \Big | n_{v}^{-1} \sum_{i=1}^{n_v} \Leg_j \big(U_{\ell i} \big) \Big|^2,  \eeq
where $\Leg_j$ denotes the $j$th orthonormalized shifted Legendre polynomial on unit interval. The $\qDIV$ statistic measures the distance between Uniform$[0,1]$ and the sample distribution of $U_{\ell}$. We have used $m=6$ in all our examples. 
One can also compute pvalues using the $\chi^2_m$ null distribution of $\qDIV$. For more details see \cite{Deep18DMT} and \cite{DeepLPKsample19}.

\vskip.2em

~Step 4. It is also interesting to note that one can \textit{directly} compute these generalized quantile-residuals from the contrast distribution function, since (by substituting $F_0(t)=u)$:
\[
F_{Y|X=x}(y)\,= \int _{-\infty}^y f_0(t) \,d_x(F_0(t))\dd t\,=\int_0^{F_0(y)} d_x(u) \dd u \,= \,D_x(F_0(y)).~~
\vspace{-.15em}
\]

For the butterfly data $\qDIV$ values (with p-values in the parentheses) for knn, random forest, and gbm respectively are $0.0466 \,(0.82)$, $1.12 \,(1.4\times10^{-10})$, and $0.386\, (0.0026)$. These statistic values can be used as a general measure of ``lack-of-fit'' for comparing (ranking or even tuning hyperparameters) the performances of different ML-engine based UPMs.

%accurately capture the underlying stochastic mechanism..

%%%%%%%%%%%%%%%%%%%%%%%%%

\subsection{Sharpening Black-box Generators}
Imagine we are given some black-box generative model that can simulate $y$-samples for a given $x=(x_1,\ldots,x_p)$:
\[\texttt{MODEL}(x)\, ~\mapsto \,~\{\wty^x_1,\ldots, \wty^x_s \}.~~~\]
Without having any information on the particular \texttt{MODEL}\footnote{This could be \textit{any} arbitrarily complex computational program, e.g., deep learning method with millions of parameters, an ensemble of $100$ machine learning models, or the method proposed in \cite{FriedmanCtree2019}.}, our goal is to: (i) check whether the samples are truly generated from the underlying conditional distribution $f_{Y|X=x}(y)$; (ii) if not, explain `why' by displaying the associated $\whd_x$; and finally, (iii) sharpen the given imperfect samples to make them compatible with the underlying stochastic data generating mechanism. 

\begin{figure}[h]
 \centering
    \begin{subfigure}[t]{.45\linewidth}
    \centering
    \caption{}
    \vskip.3em
\includegraphics[width=\linewidth,trim=1cm 1cm 1cm 1cm]{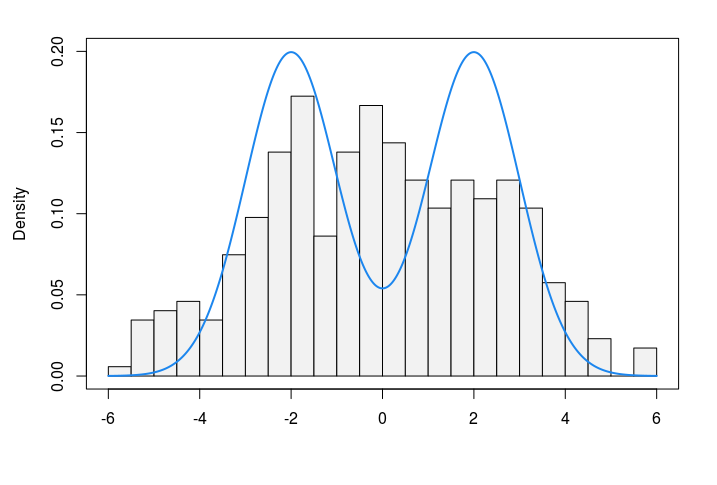}\\[2.5em]
    \end{subfigure}~~~
 \begin{subfigure}[t]{.45\linewidth}
    \centering
    \caption{}
     \vskip.2em
    \includegraphics[width=\linewidth,trim=1cm 1cm 1cm 1cm]{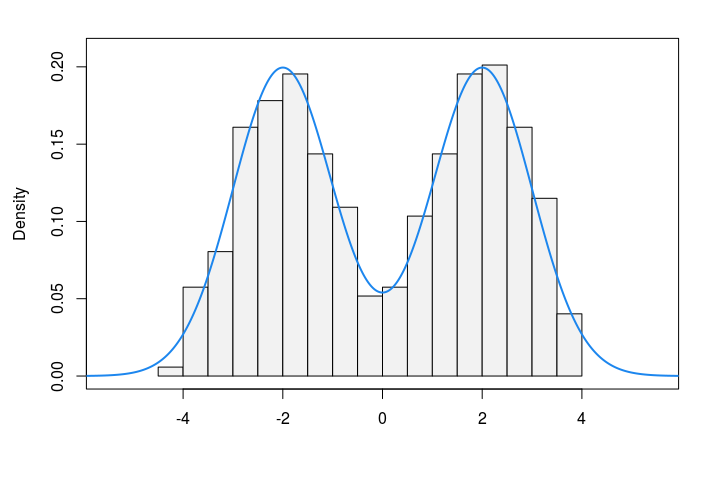}
     \end{subfigure}
\vskip.25em
\caption{Butterfly data: the histogram at left shows the $y_i$'s with $0 \le x_i\le 4$. On the right, we display the histogram of $d$-sharpened samples, which shows remarkable agreement with the underlying true conditional density (the blue line).} \label{fig:dsharp}
\end{figure}

\vskip.3em
{\bf Butterfly data}. To illustrate the method, let us consider that we are given all the $y_i$'s for which the $x_i$ value is between $0$ and $4$. The histogram of these samples is shown in the left panel of Fig. \ref{fig:dsharp}.  The blue curve shows the true $f_{Y|X=2}(y)$. Starting from these \textit{weak} conditional samples, our goal is to refine it so that they comply with the underlying law at $x=2$. 
\vskip.15em
~Step 1. Choose the empirical distribution $\widecheck{f}_{y|x}$ of $\{\wty^x_1,\ldots, \wty^x_s \}$ be the pivot; see Fig. \ref{fig:dsharp}(a).
\vskip.15em
~Step 2. Randomly generate one sample $\wty_{\mbox{*}}$ from $\widecheck{f}_{y|x}$ and $U$ from Uniform$[0, 1]$.
\vskip.15em
~Step 3. Let $\widecheck{F}_{y|x}$ denotes the cdf of $\widecheck{f}_{y|x}$. Then, set $\widehat{y}^x=\wty_{\mbox{*}}$, if
\[d_x\big( \widecheck{F}_{y|x} ( \wty_{\mbox{*}})  \big)~>~ U \max_u d_x(u). \]
Otherwise discard it and go back to the previous step. Here $d_x(u)$ means $d_x(u;\widecheck{F}_{y|x},F_{y|x})$.
\vskip.15em
~Step 4. Repeat steps 2-3 until we have $s$-sample $\{\widehat{y}^x_1,\ldots, \widehat{y}^x_s\}$, which we call \texttt{$d$-sharp} samples, since 
$d_x$ acts as a tool for sharpening the initial weak conditional samples; see Fig. \ref{fig:dsharp}(b).

%incompatibility factor.
%If would be remarkable if you can show that you can check NGBoost \citep{duan2019ngboost}: That converts Gradient Boosting into a conditional distribution.

%%%%%%%%%%%%%%%%%%%%%%%%%%%%%%%%%%%%%%%%%%%
%\section{ML-Powered Statistical Learning} 

%%%%%%%%%%%%%%%%%%%%%%%%%%%%%%%%%%%%%%%%%%%
\section{Some Connections and Unified Interpretation} \label{sec:unif}
\begin{quote}
    \textit{We should teach in our introductory courses that one meaning of statistics is ``statistical data modeling done in a systematic way''}~ \citep{parzen01}
    \vspace{-.3em}
\end{quote}
%A striking aspect of this our technique is, besides its level of generality..The new mechanics of data modeling that is presented here..
The theoretical constructs of our integrated learning framework are not based on an isolated ``trick,''  but are fundamental to statistics. Our formalism and principles of data modeling unify many traditional and modern statistical methods. In this section, we will report findings to highlight this fact.

% can be viewed as different instances of our general theory

%unified way to understand many different facets of statistics..

%Reveals some deeper statistical unity among different underneath there exists some deeper statistical theory

%not an artificially constructed ``cosmetic'' theory.We systematically derive several , which highlights its unifying role.

%Our formulation brings some delightful unity
%a unified framework to synthesize
%%%%%%%%%%%%%%%%%%%%%%%%%%%%%%%%%%%%%%%%%%%

\subsection{K-sample Comparison Problems}
Given random samples from $k$ distributions $F_i(y),$ $i=1,\ldots,k$ we seek to test the following null hypothesis:
\beq \label{eq:h0}
H_0:~ F_1(y) = \cdots = F_k(y) ~~\text{for all}~ y,~~~
\eeq
with the alternative hypothesis that at least two distributions are different. Before attacking this general problem of equality of distributions, we start with a simpler problem of testing the equality of means. The Kruskal–Wallis statistic provides a nonparametric test for it.

\begin{figure}[t]
 \centering
    \begin{subfigure}[t]{.482\linewidth}
    \centering
    \caption{}
    \vskip.3em
\includegraphics[width=\linewidth,trim=.5cm .5cm .5cm .5cm]{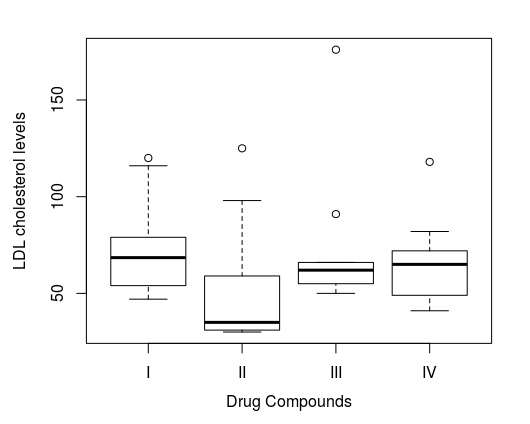}\\[2.5em]
    \end{subfigure}~~
 \begin{subfigure}[t]{.474\linewidth}
    \centering
    \caption{}
     \vskip.2em
    \includegraphics[width=\linewidth,trim=0cm .25cm .25cm 0cm]{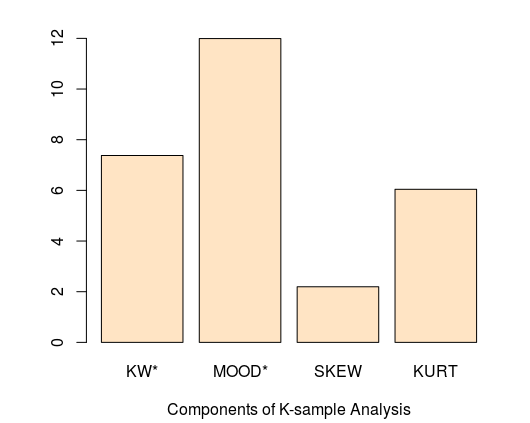}
     \end{subfigure}
\vskip.25em
\caption{LDL cholesterol data. (a) shows the boxplots of cholesterol values over four different drug components; (b) this is essentially the stretched HCA-plot. $R_j^2$ were multiplied by the sample size. The asterisk-sign `*' indicates the associated test statistic has p-value $<0.10$. In R, \texttt{kruskal.test}$(y\sim x)$ yields the value $7.18$, which is practically same as ours using \eqref{eq:kws}.} \label{fig:choleHCA}
\end{figure}

\textit{Definition and notation}. We have $k$ mutually independent random samples with sizes $n_1,\ldots n_k$ with the combined sample size $N$; $R_i$ denotes the rank of $Y_i$ in the pooled samples; $G_\ell$ is the set of indices for the $\ell$-th group. The Kruskal–Wallis test statistic is defined as:
\beq  \label{eq:kw}
{\rm KW}~=~\dfrac{12}{N(N+1)} \sum_{\ell=1}^k n_\ell \Big\{  \frac{1}{n_\ell} \sum_{i \in G_\ell} R_i - \frac{N+1}{2} \Big\}^2.
\eeq
One may go one step further and ask how to test the equality of scale parameters (variability over different groups), which can be tested using Mood statistic, defined as 
\beq \label{eq:mood}
{\rm Mood}~=~\dfrac{180}{N(N+1)(N^2-4)} \sum_{\ell=1}^k n_\ell\Big\{ \frac{1}{n_\ell} \sum_{i \in G_\ell} \big(R_i -   \frac{N+1}{2}\big)^2 - \frac{N^2-1}{12}     \Big\}^2    
\eeq
For testing high-order differences like skewness (not to mention kurtosis, which will take almost two full lines to write), the formula of the test statistic gets even more complicated.\footnote{ If you are a machine learning practitioner and concerned that these are too abstract and complicated, hang in there--don't give up. We will soon see they can be represented in a  beautiful and compact manner using our modern notation, which can be implemented in one line, without the burden of memorizing any formula.} 
{\small\beq \label{eq:skew}
{\rm SKEW}=\frac{7}{N(N+1)(N^2-4)(N^2-9)}\sum_{\ell=1}^k n_\ell\Bigg\{    \frac{20}{n_\ell}  \sum_{i \in G_\ell} \big(R_i -   \frac{N+1}{2}\big)^3  - \dfrac{3N^2-7}{n_\ell} \sum_{i \in G_\ell} \big( R_i -  \frac{N+1}{2}\big)\Bigg\}^2
\eeq
\vskip.025em
}
A practitioner at this point might wonder how to make sense of these monstrous formulae? Is any intuitive or systematic derivation possible? The answer is yes. We just have to look at the problem from a different perspective. A modern introduction to the $k$-sample comparison problem is given below.
\vskip.15em
~Step 1. View it as a $(X,Y)$ problem: Define $X$ to be the group-index variable taking values $\{1,\ldots, k\}$. We now have $\{(x_i,y_i)\}_{i=1}^N$ data, which can be displayed as a scatter plot.
\vskip.35em
~Step 2. Reformulate $H_0$: The $k$-sample null-hypothesis of equality of distributions can now be viewed as a test of the homogeneity problem (see Section \ref{sec:homog}) $f_{Y|X=\ell}(y)=f_Y(y)$ for $\ell=1,\ldots,k$. Now since
\beq f_{Y|X=\ell}(y)\,=\,f_Y(y) \Big\{ 1\,+\sum_j \LP_{j|\ell} T_j(y;F_Y) \Big\},\eeq
the original $k$-sample hypothesis \eqref{eq:h0} can alternatively be rewritten as 
\[H'_0:~ \LP_{j|\ell}=0, ~~\text{for all}~ j.~~~\]
Hence, intuitively, it makes sense to apply the HCA diagnostic test of section  \ref{sec:homog} on the $(x,y)$ data. So what happens, if we dare to take the leap of faith? We get almost all of the known results of $k$-sample modeling in one-shot.\footnote{After all, ``the man who perpetually hesitates accomplishes nothing; It is the man who dares who wins.''} To justify the statement, we start with the following theorem.
\begin{thm} \label{thm:kw}
The Kruskal–Wallis statistics can be expressed as 
\beq  \label{eq:kws}
~{\rm KW}\,=\,\sum_{\ell=1}^k n_\ell \,\big| \tLP_{1|\ell} \big|^2\,=\,N R_1^2.~~~~\text{\textup{(change in location)}}
\eeq
where the first-order LP-multiple correlation $R_1^2$ is defined in the step 1 of Section \ref{sec:homog}.
\end{thm}
The proof is in Appendix A.3. The beauty of this result is that it seamlessly generalizes to high-order cases\footnote{The agreement of our general theory with the $k$-sample results may appear to be a `miraculous coincidence' but in reality, this actually reveals the fundamental nature of it.} (proofs are not difficult and are left to the reader): 
\beas \mbox{MOOD}&=&\sum_{\ell=1}^k n_\ell \,\big| \tLP_{2|\ell} \big|^2\,=\,N R_2^2;~~~~\text{(change in scale)}\\
\mbox{SKEW}&=& \sum_{\ell=1}^k n_\ell \,\big| \tLP_{3|\ell} \big|^2\,=\,N R_3^2;~~~~\text{(change in skewness)}\\
\mbox{KURT}&=& \sum_{\ell=1}^k n_\ell \,\big| \tLP_{4|\ell} \big|^2\,=\,N R_4^2,~~~~\text{(change in tail)}
\eeas
and so on. The practical benefit of these results is that they provide a systematic and comprehensive $k$-sample analysis program that is radically simple to implement, since LP-multiple correlations $R_j^2$ can be computed in a single line R-code: \texttt{summary}(\texttt{lm}($T_j(y) \sim T_X$)\$\texttt{r.squared}.
\vskip.2em
{\bf Example 6.} LDL cholesterol of Quail \citep{hettmansperger2010book}. This is a randomized experiment investigating LDL (low-density lipoprotein) cholesterol in quails; $N=39$ quails were randomly assigned to $k=4$ diets for a specific period of time, where each diet mixed with a different drug compound. The boxplot is shown in Fig. \ref{fig:choleHCA}. The scientific question is: whether or not different drug compounds change the  LDL cholesterol levels. The right panel of Fig. \ref{fig:choleHCA} shows the re-scaled HCA-plot, where we multiply the $j$th LP-multiple-correlations $R_j^2$ by the sample size $N$. It shows that the distribution of LDL cholesterol levels is changing in scale and location.

 %evels in quails exposed to four different diets.

%They were fed over a specific period of time a special diet mixed with a drug compound. Quails were fed a same diet but mixed with different drug compounds over a certain period. A design was implemented. 

\begin{rem}
For this data, the classical anova F-test yields p-value $0.345$, thus fails to detect any significant location differences--clearly contradicting the boxplots. Why has the F-test failed? There are two primary reasons for that: (i) the highly non-Gaussian nature of the distribution of $Y$ along with a few outlying values were enough to kill the parametric F-test; and (ii) the assumption of equal variance is also seems inaccurate. The takeaway: we need \textit{both} robustness and efficiency to reliably detect high-order distributional differences.     
\end{rem}

%a consistent and coherent theory of statistical modeling.

%We believe these are just a tip of the iceberg, more and more statistical methods can be accommodated under this general framework of data modeling. Future work will directed towards this.

%%%%%%%%%%%%%%%%%%%%%%%%%%%%%%%%%%%%%%%%%%%

\subsection{Probabilistic Index Model} 
Do baseball players gain weight as they get older? Can a larger dose of an antipsychotic drug reduce the severity of depression (after adjusting for the confounders)? Do working-class Belgian families spend less of their income on food, proportionately, as they come to make more money? All of these problems are concerned with modeling the \textit{comparison probability regression}:
\beq \label{eq:cpr}
{\rm CPR}(x)~=~\Pr(Y \le Y_{|x}),
\eeq
as a function of $x$, instead of as a usual conditional mean regression. {\rm CPR}$(x)$ encodes the probability that a randomly selected subject from the population with $X=x$ has a higher response value than a randomly selected subject from the full data. This concept was pioneered by  Olivier Thas and his collaborators under the name of probabilistic index model; see \cite{thas2012PIM}. To understand how \eqref{eq:cpr} is related to our framework, we start with the following integral representation:
\beq \label{eq:lpim0}
\Pr(Y \le Y_{|x})~=~\int F_Y(y) \dd F_{Y|X=x}(y).\eeq
This leads to the following important result. 
\begin{thm} \label{thm:lpim}
$\Pr(Y \le Y_{|x})$ can be expressed as a regression of probability integral transform $F_Y(Y)$ on $X$:
\beq \label{eq:lpim}
\Pr(Y \le Y_{|x})\,=\Ex[F_Y(Y)|X=x]\,=\, \frac{1}{\sqrt{12}}\LP_{1|x}\,+\,1/2.
\eeq
\end{thm}
The first equality follows from \eqref{eq:lpim0}, and the second equality follows from supplementary equation  \eqref{eq:leg1}. The representation \eqref{eq:lpim} justifies the name comparison probability regression, which is simply a standardized 1st-order LP-regression function. As a consequence of Theorem \ref{thm:lpim}, the whole procedure can be implemented in one line of R-code: \texttt{lm}$(T_1(y)\sim T_X)$.

\begin{figure}[t]
 \centering
    \begin{subfigure}[t]{.48\linewidth}
    \centering
    \caption{}
    \vskip.3em
\includegraphics[width=\linewidth,trim=.5cm .5cm .5cm .5cm]{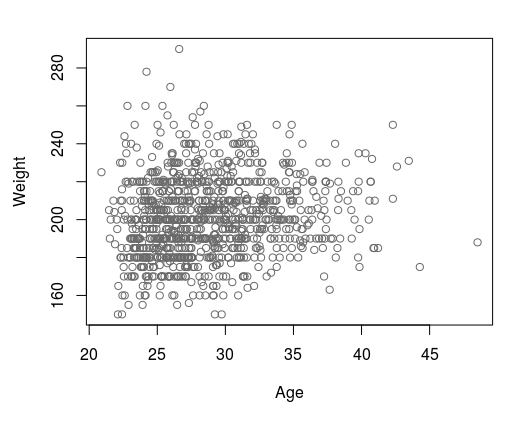}\\[2.5em]
    \end{subfigure}~~
 \begin{subfigure}[t]{.48\linewidth}
    \centering
    \caption{}
     \vskip.2em
    \includegraphics[width=\linewidth,trim=0cm .5cm .5cm .25cm]{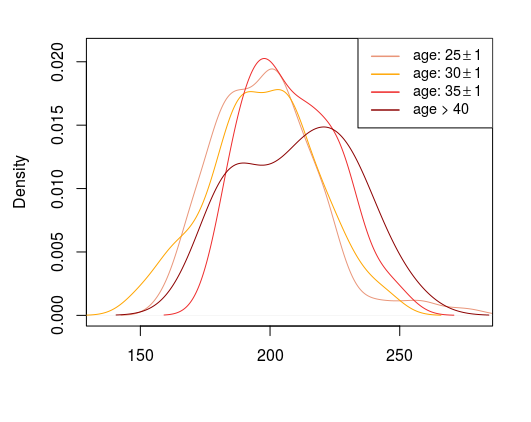}
     \end{subfigure}
\vskip.1em
\caption{(color) Baseball data. (a) Scatter plot of age versus weight of $N=1015$ major-league baseball players; (b) the density of weights for four different age groups.} \label{fig:baseball}
\end{figure}

\vskip.3em
{\bf Example 7}. \textit{Baseball data} \citep{matloff2017MLbook}. Fig. \ref{fig:baseball} shows the age verses weight scatter plot of $N=1015$ major-league baseball players. We would like to investigate whether baseball players gain weight as they age. In other words, our task is to check whether $\Pr(Y \le Y_{|x})$ increases as a function of age or not. We proceed as follows:

~~Step 1. Compute the LP-basis functions: $T_1(y;\wtF_Y)$ for the response and $T_X=[T_1(x;\wtF_X),$ $\ldots,T_m(x;\wtF_X)]$ for the predictor variable. Here we have picked $m=4$.

~~Step 2.  (penalized) LP-regression: Perform the multiple linear regression $T_1(y;\wtF_Y)$ on $T_X$. Return the Bayesian information criterion (BIC) selected model. If no variable is selected then it supports the null hypothesis $H_0: \Pr(Y \le Y_{|x})=0.5$. For the baseball data we get: 
\beq \label{eq:bball}
\tLP_{1|x}=\Ex[T_1(y;\wtF_Y)|X=x]=0.17\,T_1(x;\wtF_X).
\eeq
There is no intercept as $\Ex[T_1(y;\wtF_Y)]=0$, by construction. Although athletes strive to keep physically fit, eq. \eqref{eq:bball} seems to suggest that the baseball players do gain some weight (in a slow linear rate) over time. Fig. \ref{fig:baseball}(b) makes this somewhat clear to understand how.

%It is also important to mention that for , we recommend $L_1$-penalized least square regression techniques \citep{stanfybook}, such as lasso or elastic net

% http://heather.cs.ucdavis.edu/draftregclass.pdf

% footnote: for with ties case the Standardization is given in..which is automatically incorporated in basis construction.  
%%%%%%%%%%%%%%%%%%%%%%%%%%%%%%%%%%%%%%5
\subsection{Shape Predictors: Generalized Feature Selection} 
When dealing with a large number of covariates, it is often important to understand which are the most \emph{predictive} for the response variable $Y$.  Currently, the most popular feature selection method is $L_1$-penalized lasso regression \citep[Ch. 3.4]{stanfybook}, which minimizes the residual sum of squares $\| y-X\beta\|^2_2$ subject to $\|\be\|_1 \le \lambda$. This will be denoted algorithmically as \texttt{lasso}$(y\sim X)$. It is important to note that, when we perform lasso-regression of $y$ on $X$, we only get `first-order' informative features--the variables that influence the conditional mean $\Ex[Y|X=x]$. Naturally, the question is: can we go beyond mean, and find the most relevant attributes to describe the \emph{shape} of the conditional distribution $f_{Y|X=x}(y)$? Can we better utilize the lasso-technique to find those ``shape predictors''?  The following is an attempt to construct such a generalized feature selection program, which is extremely easy to implement. 
The algorithmic logic proceeds as follows:

~~Step 1. Let's start by recalling our basic model: $f_{Y|X=x}(y)=f_Y(y)\cdot d_x(F_Y(y))$. The covariates $x=(x_1,\ldots,x_p)$ can influence the distribution of the response $Y$ only through $d_x \circ F_Y(y)=1 + \sum_{j}\LP_{j|x}T_j(y;F_Y)$, which is governed by the following system of equations:
\beas
\LP_{1|x}&=&\Ex[T_1(Y;F_Y)|X=x]\\
\LP_{2|x}&=&\Ex[T_2(Y;F_Y)|X=x]\\
\vdots~~~& & ~~~\vdots
\eeas
The first equation captures the dynamics of conditional mean as a function of $x$, while the second one describes how the variability (scale) changes with $x$, and so on.

~~Step 2. Thus, it is quite obvious that one can extract the $j$th order features, denoted by  
$$\bO_j\,=\,\big\{\ell: \widehat \be_\ell^{(j)} \neq 0\big \},$$
where $\widehat{\be}^{(j)}$ is the output of \texttt{lasso}$(T_j(y)\sim X)$, which efficiently computes the solution of 
\beq  
\widehat{\be}^{(j)}~=~\argmin_{\be^{(j)} \in \cR^{p}} \Big\{ \big\| T_j(y)\,-\,X \be^{(j)}\big\|^2_2  ~+~\lambda \| \be^{(j)}\|_1 \Big\}.
\eeq
Accordingly, $\bO_1$ is the collection of location-informative variables, $\bO_2$ contains all the scale-informative ones, $\bO_3$ variables are responsible for predicting change in skewness of $Y,$ etc. 

~~Step 3. To make the analysis doubly-robust and to allow for non-linearity of the features, simply substitute $X$ by the LP-transformed matrix $T_X$, as we did in Sec. \ref{sec:gALGO}. 
\begin{defn}
Let $\widehat{\be}_{\ell k}^{(j)}$ be the estimated lasso regression-coefficient for $T_k(x;\wtF_\ell)$. Then define the Omnibus Variable Importance Score (\texttt{Ovis}) for $X_\ell$ as
\beq \label{eq:ovis}
\mbox{Ovis}_\ell~=~\sum_j \sum_k \,\big|\, \widehat{\be}_{\ell k}^{(j)} \big|,^2~~{\rm for}~\ell=1,\ldots,p.
\eeq
\end{defn}

\begin{rem}[Portability]
Our procedure for finding generalized shape predictors is extremely flexible: one can easily integrate it with \emph{any} machine learning algorithm. Train the model \texttt{ML}($T_j(y) \sim X$), and plug that into a feature-importance wrapper e.g., h2o.varimp() from \texttt{h2o} R-package, varImp() from \texttt{caret} R-package, or FeatureImp() from \texttt{iml} R-package, etc.
\end{rem}

\begin{figure}[t]
    \centering
    \includegraphics[width=.8\linewidth,trim=.7cm .15cm 1cm .5cm]{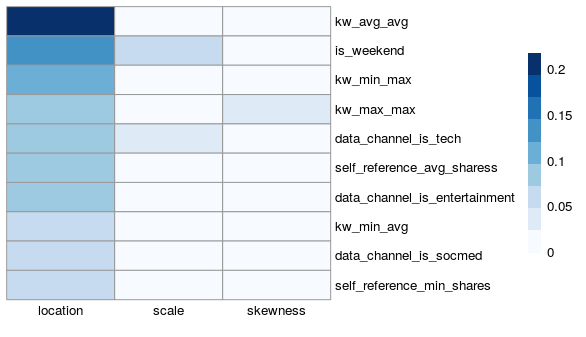}
    \caption{Online news data: Top ten generalized shape predictors are shown. They are classified based on how they impact the response distribution $f_{Y|X=x}(y)$. } %They significantly contribute in changing shape of the conditional distribution.
    \label{fig:gsp_news}
\end{figure}
{\bf Example 8.} \textit{Online news popularity data, continued}. Which attributes are most important for predicting the popularity of a given article?  More specifically,
which variables affect the changing shape of the conditional distribution? Our approach consists of four main steps: 

~~$\bullet$ First, determine which shape parameters of $f_{Y|X=x}(y)$ are changing with $x$? In Fig. \ref{fig:hcomps}, we have already seen that scale and skewness are the two principal dynamic components, in addition to location. 

~~$\bullet$ Second, to determine the variables that impact the changing location, scale, and skewness we perform the following LP-lasso regression\footnote{see Appendix A.4.4 for a related discussion on `robust lasso.'} 
\[\texttt{lasso}\big(T_j(y) ~\sim~T_1(x_1) + \cdots+T_1(x_p)\big), ~~\text{for}~j=1,2,3,\]
where $T_k(x_\ell):=T_k(x;\wtF_\ell)$. Store the selected features in the set  $\bO_j$.

~~$\bullet$ Third, we summarize the findings by ranking the variables using the \texttt{Ovis} index; here we use $k=1$ and $j=3$ in the formula  \eqref{eq:ovis}. The top $10$ features are displayed in Fig. \ref{fig:gsp_news}, which shows the \emph{nature} of contributions of different variables. For example, the variable `\texttt{is\_weekend}' (whether the news article was published on the weekend or not) plays a dual role of being both a location and scale informant. Consequently, our multi-layered robust method provides a refined understanding of the ``which and how'' aspects of feature selection.
%The Fig. \ref{fig:gsp_news} is showing top 10 influencing variables for $\bO_1$, $\bO_2$ and $\bO_3$ .

~~$\bullet$ Finally, one can perform \emph{targeted} feature screening. An investigator can specifically query which variables are mostly responsible for heteroscedasticity, or change in symmetry, etc. Table \ref{tab:my_label} shows the top five variables in each category: location, scale, and skewness.  These insights will ultimately help the applied researchers to better understand the \emph{nature} of the association between response $Y$ and $X=(X_1,\ldots,X_p)$.

%Fig. \ref{fig:gsp_news} shows the $\bO_1$, $\bO_2$ and $\bO_3$ variables, which affect the conditional mean and variance. As we can see there are quite a number of variables that are second-order informative variables...we would have ..vanilla lasso would have missed them.

%$9$ variables are selected as a mean feature. $16$ as a scale-feature. And $4$ are common in both sets: 

\begin{table}[t]
    \centering
    \begin{tabularx}{.96\linewidth}{YYY}
    \toprule
     $\bO_1$ (location) &  $\bO_2$ (scale) &  $\bO_3$ (skewness)\\
    \midrule
    kw\_avg\_avg &is\_weekend&kw\_max\_max            \\ 
    is\_weekend & LDA\_04 & LDA\_03 \\
    kw\_min\_max &timedelta & data\_channel\_is\_socmed\\
    self\_reference\_avg\_shares  &kw\_min\_min & data\_channel\_is\_tech\\
    data\_channel\_is\_tech  &LDA\_03 & weekday\_is\_saturday\\
    \bottomrule
    \end{tabularx}
    \caption{Online news data: Top five variables for each of the three categories are shown. Investigators can use this tool to identify different sources of heterogeneity.}
    \label{tab:my_label}
    \vspace{-.4em}
\end{table}

%deduced systematically from some underlying general principle.
%%%%%%%%%%%%%%%%%%%%%%%%%%%%%%%%%%%
\subsection{The XYZ Problem: Distributional Impact Analysis}
In application fields such as healthcare, economics, and social sciences, data often arise in \texttt{XYZ} format, where $Z$ is the binary treatment variable, $Y$ is the response variable, and $X$ is the (possibly large) collection of covariates. One such problem is discussed below.

{\bf Example 9}. \textit{Rosner's FEV data} \citep{rosner95book}. Table \ref{tab:fev_snapshot} describes the data. 
The main interest lies in understanding the impact of smoking on the \emph{distribution} of FEV as a function of age.  This is important for individualized custom-tailored decision-making, where a treatment might be \emph{locally}-effective for a sub-population (characterized by certain values of $x$) without being globally effective for the whole heterogeneous population. 
\begin{table}[h]
\vskip.64em
    \centering
    \begin{tabularx}{.5\linewidth}{YYY}
    \toprule
         $X$ &  $Y$ & $Z$\\[-.3em]
        (Age)&(FEV)& (Smoking)\\
        \midrule
        9 & 1.70 & 0  \\[.2em]
        8 & 1.72 & 0  \\[.2em]
        $\vdots$ & $\vdots$ & $\vdots$\\[.2em]    
        15 & 3.73 & 1 \\[.2em]
        18 & 2.85 & 0 \\[.1em]
    \bottomrule
    \end{tabularx}
    \vskip.25em
    \caption{We are given $N=654$ observations on youths aged $3$ to $19$ from East Boston recorded during 1970's. The outcome variable $Y$ is forced expiratory volume in 1 second--a measure of lung capacity. $Z$ is $1/0$ binary treatment (smoker: yes or no) indicator variable.}
    \vspace{-.54em}
    \label{tab:fev_snapshot}
\end{table}

Our model for conditional distribution of $Y$ given $X=x$ and $Z=z$ is 
\beq \label{eq:xyz}
f_{Y|x,z}(y)=f_Y(y) \big\{ 1\,+\,\sum_j \LP_{j|x,z} T_j(y;F_Y)\big\}.\eeq
Distributional impact analysis aims to quantify (how much) and characterize (in which ways) the differences between $f_{Y|x,1}(y)$ and $f_{Y|x,0}(y)$. To develop a measure of distributional impact, first note that from equation \eqref{eq:xyz}, both $f_{Y|x,1}$ and $f_{Y|x,0}$ have the same pivot $f_Y(y)$. Thus the deviation between them directly depends on the distance between the contrast densities $d_{x,0}$ and $d_{x,1}$. 
\begin{thm} \label{thm:dif} The $L^2$ distance between the $d_{x,1}(u)$ and $d_{x,0}(u)$ can be expressed as $\ell^2$ distance between the corresponding LP-Fourier coefficients.
\beq \label{eq:ddif}
\int_0^1\Big(d_{x,0}(u) - d_{x,1}(u) \Big)^2\dd u~=~\sum_{j} \big| \LP_{j|x,0}\,-\, \LP_{j|x,1}\big|^2.
\eeq
\end{thm}
The proof is in Appendix A.3. This result motivated us to define the following measure. 
\begin{defn}
Define the distributional impact function for a given $X=x$:
\beq \label{eq:DIF}
{\rm DIF}(Y,Z \mid X=x)\,= \sum_{j} \big| \LP_{j|x,0}\,-\, \LP_{j|x,1}\big|^2.~~
\eeq
The components of the \texttt{DIF} capture the amount of change in the location, scale, etc. For example, the location difference can be estimated from the first-order LP-difference statistic: $\LP_{1|x,0}- \LP_{1|x,1}$.
\vspace{-.4em}
\end{defn}
\begin{figure}[t]
    \centering
    \vspace{-1em}
    \includegraphics[width=.94\linewidth,trim=.5cm 0cm .5cm .5cm]{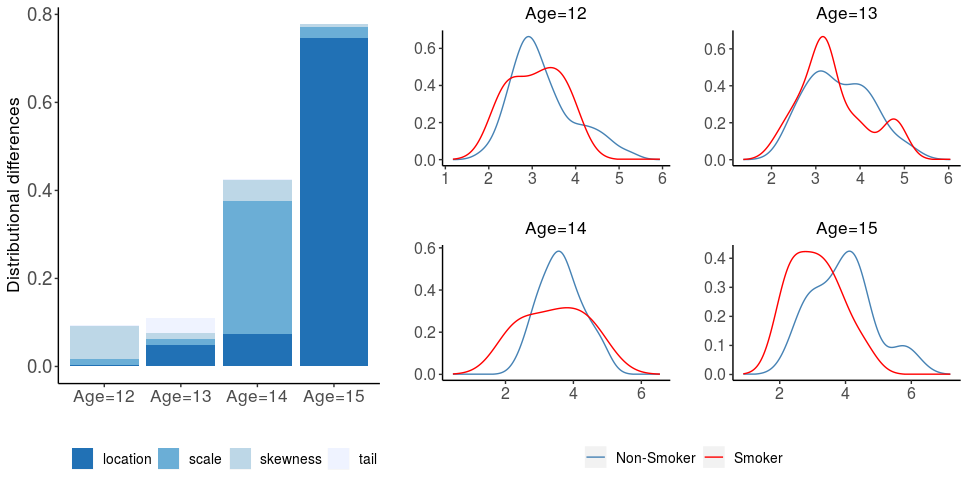}
    \caption{Rosner's FEV data. Our analysis reveals the heterogeneous nature of the impact of smoking on the distribution of FEV, across different age groups. The findings of \texttt{DIF} barplot excellently corroborated by the  kernel density estimates of the right panel.}
    \label{fig:fev_DIF_cden}
    \vspace{-.64em}
\end{figure}
\vspace{-.4em}
We are now ready to apply this procedure on the FEV data. To estimate the LP-coefficients in \eqref{eq:xyz} we apply our \texttt{UPM} framework (algorithm of Section 2.3.2) with gradient-boosting machine as the base learner. We chose gbm, as it efficiently computes the interactions\footnote{
Treatment$\times$covariates interactions allow the treatment effect to vary among individuals with covariates.} between the treatment and covariates. The left panel of Fig. \ref{fig:fev_DIF_cden} shows the \texttt{DIF} statistic for ages $12$ to $15$. The stunning fact about this graph is that it simultaneously answers \emph{which} $x$-groups are most impacted by the treatment and \emph{how} they are impacted. For example, at age $14$ variability is the dominant mode of difference, whereas at age $15$, location is the main source of difference. Overall, the \texttt{DIF} barplot does a modest job of capturing the heterogeneous impact of teenage smoking on lung function across different age groups. 
\begin{rem}[Connection with causal inference]
For randomized experiments and clinical trials our \texttt{DIF} statistic can be interpreted as a `distributional' treatment effect\footnote{The topic of `distributional' treatment effect (as opposed to `mean' effects) carries a major significance in modern econometrics \citep{bitler2006,imbens2009recent,banerjee2015miracle}, where it is known that the impact of an intervention on the outcome distribution can be highly heterogeneous--somewhat similar to what we have already seen in Fig. \ref{fig:fev_DIF_cden}. The good news is: \texttt{DIF}($Y,Z|X$) provides a systematic way to characterize the impact of a treatment $Z$ across the entire distribution of $Y$ as a function of the covariates $X$.}, one whose impact vary across sub-populations. For observational studies, one needs to assume the so-called `unconfoundedness' assumption to attach a causal interpretation.
%For observational studies, one needs to impose the so-called `unconfoundedness' assumption to attach a causal interpretation. 
\vspace{-.5em}
\end{rem}
% But we will not make such a claim for this dataset, which is a observational study..requires additional assumptions to attach the ...as we are not able to verify it..we will refrain from this.. the assumption is untestable

% [Assumptions] page 4: https://arxiv.org/pdf/1706.09523.pdf

% Ignorability/ all confounders measured.. it means that we have conditioned on all
% confounders (pre-treatment variables that predict both
% treatment and outcome)
% 

%%%%%%%%%%%%%%%%%%%%%%%%%%%%%%%%%%%
\subsection{Quantile Regression}
Quantile regression, pioneered by \cite{koenker1978}, has become a pervasive tool in a host of real-world application areas. Despite the great progress made in the last 40 years \citep{koenker2017quantile}, some practical challenges still remain: how to develop a \emph{nonparametric} quantile regression method that can scale to high-dimensional problems? How to efficiently incorporate nonlinearity? How to ensure that the estimated conditional quantile curves will not cross each other? Our statistical learning theory can offer some realistic solutions to these frontier problems of quantile regression. But before getting into that, let us start with a slightly broader context. Our ``interlinked'' statistical learning architecture (as depicted in Fig. \ref{fig:upm_flowchart}) has two notable consequences for the practice of nonparametric data modeling: 
\vskip.1em
~$\bullet$ Our technology opens up brand-new ways of building ``ML-powered'' statistical models that are simultaneously flexible and scalable\footnote{By taking advantage of high-performance specialized ML-hardware and software architecture.} for large($n,p$) problems, where classical nonparametric methods are not a viable option. 
\vskip.1em

%Classical  quickly become intractable for high-dimensional data. This is where our technology comes in: it .  

%which were beyond reach for conventional statistical methods
%power the advancement of realistic statistical modeling .

%Every alternative month we see a new avatar of an old algorithm. 
%Some even say that a new ML-algorithm `on average' has a shorter shelf-life than the lifespan of fruit flies. 
~$\bullet$ Our UPM learning architecture is designed using a high-level universal language that is agnostic to the specific ML method, to ensure the portability and durability of the technology.\footnote{It is important to keep in mind that ML algorithms are growing at a staggeringly fast rate. In these circumstances, it is nor practical to develop `retail' strategies on a case by case basis for each ML-procedure. A practical way out is to build a unified learning ecosystem.}

\vskip.1em
To perform UPM-based quantile regression, simply extract the appropriate percentiles from the conditional density estimate $\hf_{Y|X=x}(y)$. The beauty of our method is that we don't have to develop theories specialized to a certain kind of machine learning method like random forest \citep{meinshausen2006quantile,athey2019grf} or neural network \citep{cannon2011quantile}--it's all done using a single generic procedure; see Appendix A.4.3 and Fig. \ref{fig:app_quantreg} for more details.

%computationally tractable for modern-day complex datasets of enormous size
 
\begin{figure}[t]
    \centering
    \vskip.65em
    \includegraphics[width=.6\linewidth,trim=2cm .5cm 2cm 1.5cm]{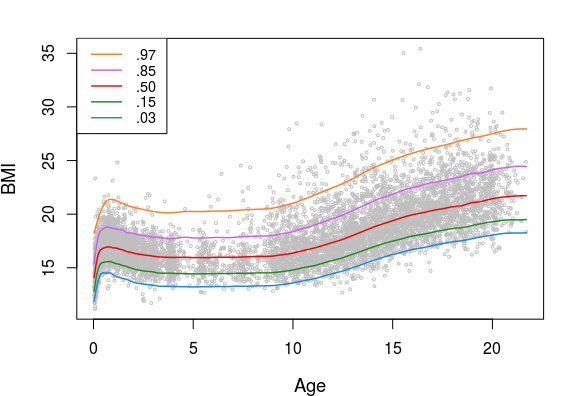}
    \vskip1.5em
    \caption{Dutch boys data: The UPM-estimated conditional quantile curves. Our method yields non-crossing quantile curves, as they are extracted from the conditional distributions.} 
    \label{fig:auto}
\end{figure}

\vskip.1em
{\bf Example 10}. \textit{Dutch Boys data} \citep{fredriks2000body}. This dataset is a part of the Fourth Dutch Growth Study, which comprised of observations on age and BMI of $N=7294$ Dutch boys. The goal is to estimate age-specific reference growth curves based on conditional quantile function $\whQ_{Y|X=x}(u)$. Fig \ref{fig:auto} displays our result where we have used \texttt{lasso} (as implemented in the glmnet R-package) as the base learner inside \texttt{UPM} engine. Following the current World Health Organisation (WHO) recommendation, we have used $u=(.03, .15, .50, .85, .97)$ for constructing the cross-sectional reference growth charts. 

\begin{rem}[Personalized reference interval prediction]
The topic of individualized referencing carries special significance in the era of precision medicine. To accurately infer each individual's reference curves (describing the  \emph{normal} range of the outcome $y$ given $x$) medical researchers are now incorporating more and more variables in the form of genetic and environmental factors.  Thus it will be of great practical value to have a method that can tackle a large number of covariates, and do so in a completely nonparametric and nonlinear fashion.
\vspace{-.5em}
\end{rem}
%%%%%%%%%%%%%%%%%%%
%%%%%%%%%%%%%%%%%%%%%%%%%%%%%%%%%%%%%%%%%%
\subsection{Prediction Interval Estimator} \label{sec:PI}
One way to summarize the conditional density is through prediction interval (PI), which provides a concise view of the most likely values of the response variable\footnote{Knowing the range of values of $Y$ as opposed to a single-point estimate, helps decision-makers choose proper action by carefully evaluating the degree of uncertainty (length of the PIs).}. For a specified level $\al$, the goal is construct an interval that covers no less than ($1-\al$) of the probability mass of $f_{Y|X=x}(y)$. How to construct such PIs? We discuss three different procedures. The real challenge is to make it as narrow as possible, while maintaining the desired coverage.

%distribution estimator to interval estimator.

%%%%%%%%%%%%%%%
~$\bullet$  \textit{Quantile-based PI}. As noted by \cite{meinshausen2006quantile}, one can use  quantile regression to build PIs. A $100(1-\al)$\%  prediction interval
for $Y$ given $X=x$ can be estimated as
\beq \label{eq:qpi}
{\rm qPI}_{1-\al}(x)~=~\left[Q_{Y|X=x}(\al/2),~ Q_{Y|X=x}(1-\al/2)\right].
\eeq
Naturally, the width (precision) of the quantile-PIs vary over the covariate space. 

%%%%%%%%%%%%%%%
~$\bullet$  \textit{Standard-error-based PI}. If we are brave enough to assume some parametric form of the conditional density, then we can greatly simplify the expression for the PIs.  For example, under Gaussianity assumption, one can express \eqref{eq:qpi} in the following compact form, just involving conditional mean $\mu(x)$ and standard-deviation:
\beq \label{eq:gpi}
{\rm gPI}_{1-\al}(x)~=~\left[ \mu(x) \pm z^{(\al/2)} \si_{|x} \right]~~~~~~~~~~~~~~~~~~~~~~
\eeq
where $\Phi^{-1}(\al)=z^{(\al)}$, $ \Var(Y|X=x)=\si^2_{|x}$, and `g' denotes the Gaussian-PI.

\begin{figure}[t]
    \centering
    \begin{subfigure}{.51\linewidth}
    \caption{Butterfly data} 
        \includegraphics[width=\linewidth,trim=.5cm .75cm .1cm .5cm]{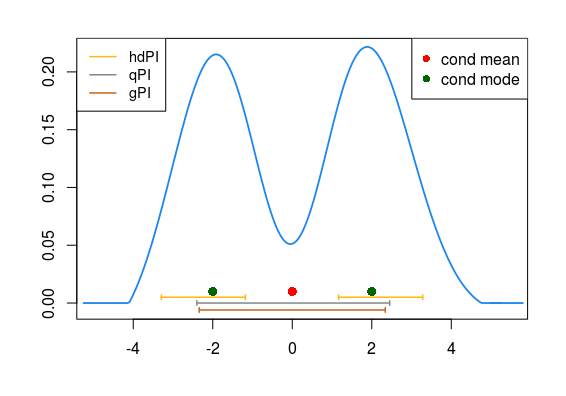}
    \end{subfigure}~~
    \begin{subfigure}{.51\linewidth}
    \caption{1982 Dodge Rampage} 
        \includegraphics[width=\linewidth,trim=.5cm .75cm .2cm .5cm]{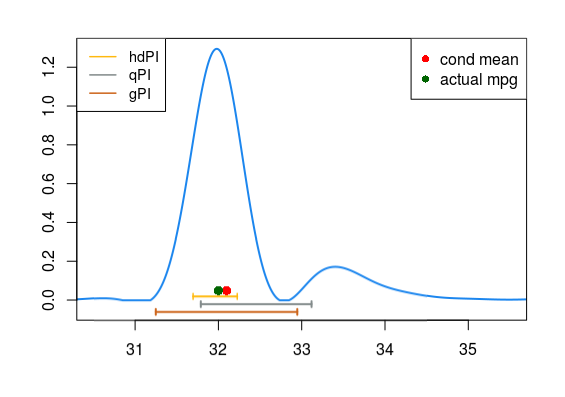}
    \end{subfigure}
    \vskip.25em
    \caption{(color) Left: Butterfly data. Three different prediction intervals for $Y|X=2$ along with the conditional mode and mean estimates. All are 68\% PIs, equivalent to $\pm 1 \sigma$ uncertainty. Right: the curious case of 1982 Dodge Rampage. Two notable things: the shape of the conditional density estimate and the contrasting lengths of different PIs. \texttt{hdPI}s tend to produce narrow, more precise prediction intervals while ensuring the desired coverage. \texttt{qPI} seems to be little misaligned, due to the long right-tail.} 
    \label{fig:PI}
\end{figure}

%%%%%%%%%%%%%%%
~$\bullet$  \textit{Highest-density PI}.
This is the shortest possible interval with a given coverage probability \citep{boxbook73}, defined as 
\beq \label{eq:hdpi}
{\rm hdPI}_{1-\al}(x)~=~\left\{y: f_{Y|x=x}(y) > \tau_{1-\al}\right\} \eeq
where $\tau_{1-\al}$ is the largest number satisfying 
\[\int_{y \,\in\, {\rm hdPI}_{1-\al}(x)} f_{Y|X=x}(y) \dd y = 1-\al.\]
%conditional on $X=x$. But it requires the estimation of conditional density..
Another nice property of \texttt{hdPI}s is that any point within the interval has a higher density than any other point outside it, which justifies their name. 

Based on their operation principles, these three kinds of PIs can be broadly categorized into two groups: the vertical and the horizontal approach.

%%%%%%%%%%%%%%%%%%%%%%%%%%%

{\bf Example 11} \textit{Butterfly data, continued}. All three prediction intervals are shown in Fig. \ref{fig:PI}(a). Three main conclusions: (i) The shortest among all is the highest-density PI, which take the form of disjoint subintervals--one for each local mode. (ii) The Gaussian assumption of {\rm gPI}  seems too restrictive for real-world data analysis.  (iii) Both {\rm gPI} and {\rm qPI} include a low-density valley area around $x=0$ at the cost of other, more-likely values around $x=\pm 2$. 

% better understanding of the uncertainty in forecasts..Indeed, if we predict the lower and upper quantiles of the target then we will be able to obtain a “trust region” in between which the true value is likely to belong. 

%After removing the missing entries, we have $N=392$ samples. 

{\bf Example 12}.  \textit{The Auto-MPG data}. The task is to predict the miles-per-gallon (MPG) gas consumption of an automobile based on $p=7$ features including horsepower, weight, and acceleration. The dataset was used in the 1983 American Statistical Association Exposition and is available in the UCI Machine Learning Repository. Here, we are mainly interested in predicting the MPG of a specific instance--row number $390$, which corresponds to the 1982 Dodge Rampage car. The \texttt{UPM}-predicted uncertainty distribution is displayed in Fig. \ref{fig:PI}(b). The bimodality of the estimated conditional density reflects the fact that 1982 Dodge Rampage is a \emph{hybrid} two-door car--a mix between a passenger car and a truck. The next notable thing is the length of the PIs: The standard error-based \texttt{gPI} is almost three times wider than the \texttt{hdPI}, and thus provides an overly pessimistic assessment of an accurate prediction. This also sheds some light on the apparently paradoxical fact (as noticed by \citealt{wager2014confidence}) why bootstrap standard error-based Gaussian confidence interval yields a large error bar for the Dodge Rampage example, even when the prediction was dead on.
%For unimodal and symmetric distributions
%%%%%%%%%%%%%%%%%%%%

\subsection{Modern Applied Statistics: Theory, Practice, and Pedagogy}
A theoretical research, which has no relevance to practice and teaching, is not persuasive enough to be taken as fundamental.  To us, it is very important to know whether our research was able to link (at least partially) the triad of theory-practice-teaching. In this article, through several examples, we described `the joy of systematic data analysis' derived from a general theory, which has the following implications for the practice and pedagogy of statistics:

~$\bullet$  Our theory makes it easy: to apply (arise from simplification) and see the connections (arise from unification) between different statistical methods.

%Similar philosophy was echoed by Freeman Dyson in his essay on `Birds and Frogs': https://www.ams.org/notices/200902/rtx090200212p.pdf.

~$\bullet$  Our theory provides a holistic training\footnote{For more discussion on this topic see \cite{Deep17Teaching}.} by introducing a large variety of statistical topics (embracing different cultures: algorithmic, parametric, nonparametric, information-theoretic, robust, exploratory) in an \emph{organized} manner through a common semantics.  This is extremely important to produce broad integrator foxes rather than highly specialized hedgehogs.\footnote{Readers may find it worthwhile to read Philip Tetlock's book on `Why Foxes Are Better Forecasters Than Hedgehogs,' which is based on a 20-year study with 284 experts from diverse fields, including government officials, professors, journalists, and others. As Richard Feynman said ``Science is not a specialist business.''} Ultimately, what matters is whether our theory provides a faster and easier (lazier?) way to learn the fundamentals of statistics. We believe--yes, it does.\footnote{At least it has a better chance than any other currently known \emph{global} theory of data analysis. We derive confidence from the ability of our theory to tackle statistical problems as diverse as time-series analysis \citep{D12e}, copula modeling \citep{D20copula}, empirical Bayes \citep{deep18nature}, graph-theory \citep{Deep19SGT}, multiple testing \citep{Deep18DMT, deep16LSSD}, high-dimensional data analysis \citep{DeepLPKsample19}, large-scale distributed learning \citep{deep16MetaLP}, etc. This is an ongoing and growing movement with the mission to structure the field in an understandable and effective way.}

%

%%%%%%%%%%%%%%%%%%%%%%%%%%%%%%
\section{70 Years of Statistical Machine Learning}
Breiman's 2001 paper was highly influential and hailed as a heroic effort to challenge the status quo in a ``David and Goliath'' style, where two diametrically opposite modeling cultures--machine learning and traditional parametric statistics--battled for the throne. However, it is much less well-known among practitioners (including the current generation of deep-learning engineers) that machine learning has strong historical roots in (nonparametric) Statistics. In an unpublished US Air Force School of Aviation Medicine report in 1951, two UC Berkeley Statisticians (students of Jerzy Neyman) Evelyn Fix and J.L. Hodges, Jr., introduced a non-parametric method for pattern classification, which is now called the k-nearest neighbor (knn) method. This was a groundbreaking paper that marked the beginning of statistical machine learning\footnote{Interestingly, just after that, Arthur Samuel came up with the phrase “Machine Learning” in 1952.}. Fix and Hodges's work was a precursor to the celebrated kernel-based methods  and decision tree-based algorithms, which revolutionized the world of machine learning. The time has come that we bring machine learning back to its statistical roots.\footnote{The next revolution of machine learning will need both power of computing and power of statistical thinking (and little bit of common sense).} \nocite{fix51} 
%\citep{lin2006random} \citep{silverman1989fix}

%\textit{Motivation} This research was born out of the dissatisfaction with the existing  disunity between these two communities. 

%matured over the last seven decades...integrative statistical learning framework 
%\textit{Two-in-One model}. 

\vskip.25em

The current paper challenges Breiman's ``two-culture'' model in favor of an \emph{integrated learning system} where different cultures can be ``glued'' together by some underlying deeper principles.\footnote{There is no doubt that in the coming years we will witness many more new varieties and cultures of data analysis, but in the midst of those temporary excitements, we should not forget to keep the discipline internally consistent. Diversity \emph{without} unity would be a complete mess --not healthy for our profession.} Through numerous examples, we have shown that the two cultures are not contradictory but complementary to each other, which when combined properly, can yield a more powerful learning technology that is \emph{simultaneously} flexible, scalable, and explainable. 

\vskip.25em

If there is one thing that our work has made apparent is: \emph{there's plenty of room in the middle}--tremendous opportunities lie at the boundaries between the two cultures. In this paper we have described a new theory for constructing a \emph{universal} interface between statistics and machine learning. What we have proposed is by no means the only possible or even the ``best'' theory; it is merely a provisional one. As we go forward, it will be very important to come up with new innovative theories that can reconcile and build new links between various cultures of statistical modeling. All of this is still in its nascent stages, but definitely poised for rapid progress in the coming decade.

\section*{Acknowledgement}
%This paper is written, keeping two historic occasions in mind: 
This paper celebrates the 200th anniversary of Gauss-Markov least-square regression and 70th anniversary of statistical machine learning\footnote{These two historical moments symbolize the birth of `two cultures' of data modeling--linear parametric and flexible algorithmic modeling cultures.},  by putting forward a modern unified perspective.  
% %%%%%%%%%%%%%%%
\bibliographystyle{Chicago}
\bibliography{ref-bib2}

\newpage

\renewcommand{\baselinestretch}{1.4}
\setlength{\parskip}{1.6ex}
\clearpage
\renewcommand{\theequation}{E\thesection.\arabic{equation}}
\begin{center}
{\Large {\bf 6.~~SUPPLEMENTARY APPENDIX}}\\[-.5em] % 
% with Some Examples 
\end{center}
\addcontentsline{toc}{section}{6~~~Appendix}
\subsection*{A.1. Software} 
We have developed an R Software \texttt{LPMachineLearning}\footnote{A preliminary version is available from the authors. A more upgraded one is currently under development.} to perform all the tasks outlined in the paper. We hope this software will encourage applied data scientists to apply our method for their real prediction problems.

%which is available in CRAN at https://cran.r-project.org/web/packages/LPMachineLearning.

%We provide an Rpackage,...(cite)  We now summarize the main functions and their usage for the Boston Housing data example..The package also provide functionalities for

\vskip.6em
\subsection*{A.2.~LP-Orthonormal System: Construction and Properties}
LP-basis functions are specially-designed orthonormal system of rank-polynomials. They are inherently `nonparametric,' constructed in a fully data-driven way, not pre-defined like classical polynomials. 
For a mixed (either discrete or continuous) random variable $Z$ with distribution function $F_Z$, we describe the construction of LP-basis functions.
\vskip.4em

{\bf Construction Algorithm}. A fully-automated construction of LP-orthonormal rank-polynomials of $Z$ consists of the following steps:
\vskip.3em
~~\texttt{Step 1}. Mid-Distribution Transform:
The mid-distribution function of $Z$ is defined as 
\beq 
\Fm_Z(z)=F_Z(Z)-\frac{1}{2}p_Z(z),
\eeq
where $p_Z(z)$ is the probability mass function.
\vskip.2em
~~\texttt{Step 2}. Standardizing $\Fm_Z(Z)$: The random variable $\Fm_Z(Z)$ has mean $\Ex[\Fm_Z(Z)]=.5$ and  variance $\Var[\Fm_Z(Z)]=\frac{1}{12}\big( 1- \sum_z p_Z^3(z) \big)$. Define the 1st-order LP-basis function as
\beq 
\label{eq:LP1st}
T_1(z;F_Z)~=~\dfrac{\sqrt{12}\big\{\Fm_Z(z) - 1/2\big\}}{\sqrt{1-\sum_z p_Z^3(z)}},
\eeq
\vskip.1em
~~\texttt{Step 3}. Gram-Schmidt Orthonormalization: Construct high-order polynomials by applying Gram-Schmidt orthonormalization (see Appendix A of \cite{D20copula} for a quick refresher on Gram-Schmidt process) on the set of functions of the power of $T_1(Z;F_Z)$. 
%The data-adaptive shapes of LP-basis functions.. see Supplementary S1 of \cite{DeepLPKsample19} for more details. 

\vskip.4em
{\bf Some properties}. Here we list some important properties of the They satisfy the following properties of the LP-orthonormal systems:

1) LP-bases are orthonormal with respect to the measure (weighting function) $F_Z$:
\[ \int T_j(z;F_Z) \dd F_Z =0,\,~~\text{and} \,\,\int T_j(z;F_Z) T_k(z;F_Z) \dd F_Z =\delta_{jk}.\]
For related discussions, see Remark 4 of \citet[p.\,82]{D20copula1}.

\vskip.35em
2) For real-data analysis, construct empirical LP basis (in short \texttt{eLP} basis) $\{T_j(z;\wtF_Z)\}_{j=1,2\ldots,m}$, where $m$ is strictly less than the number of unique values in the sample $\{z_1,\ldots, z_n\}$. \texttt{eLP} bases are orthonormal system of functions with respect to the discrete empirical measure $\wtF_Z$.

\vskip.15em

\begin{figure}[t]
    \centering
    \begin{subfigure}[t]{.46\linewidth}
    \caption{Weight}
        \includegraphics[width=\linewidth,trim=.5cm 0cm 0cm .5cm]{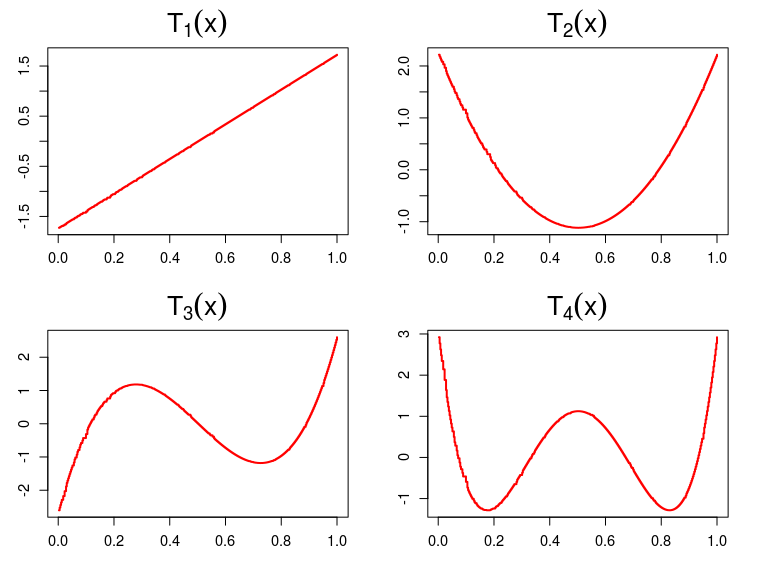}
    \end{subfigure}~
    \begin{subfigure}[t]{.46\linewidth}
    \caption{Acceleration}
        \includegraphics[width=\linewidth,trim=0cm 0cm .5cm .5cm]{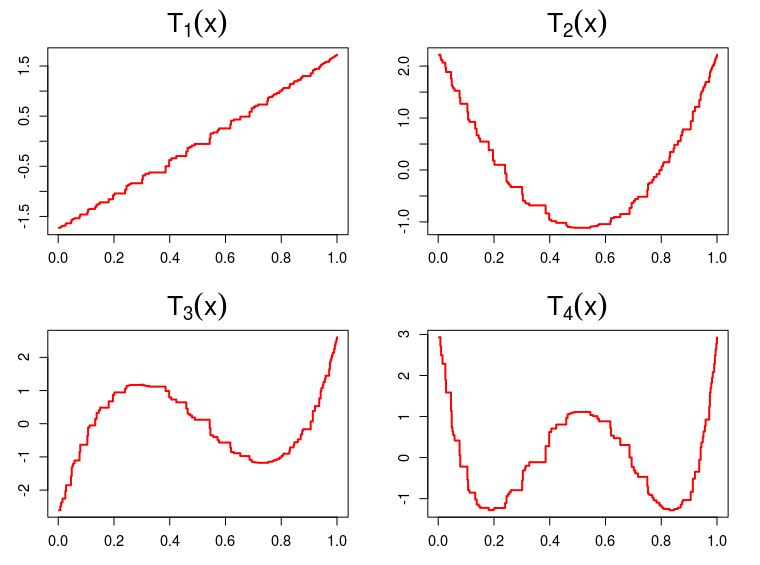}
    \end{subfigure}
    \vskip.24em
    \caption{First four LP-basis functions for the covariates (a) \texttt{weight} and (b) \texttt{acceleration} of Auto-MPG data. They are plotted with respect to $\wtF_X(x)$ over unit interval $[0,1]$, following the equation \eqref{eq:Ulp}.}
    \label{fig:app_lppoly}
    \vspace{-.25em}
\end{figure}

3) For $Z$ continuous, it is not difficult to show that the LP-bases reduces to the following universal shape:
\beq \label{aeq:leg}
T_j(z;F_Z) := \Leg_j \hspace{-.08em}\circ \,F_Z(z),\eeq 
where $\Leg_j(u)$ denotes $j$th shifted orthonormal Legendre polynomials on $[0,1]$ and `$\circ$' denotes composition of functions.\footnote{In our regression context the response variable $Y$ is continuous, thus admits this analytic LP-basis form. However, this one-to-one correspondence between LP-polynomials and Legendre polynomials of rank-transform (also know as probability integral transform) $F_Z(Z)$ is not true for discrete $Z$, i.e., for the covariates $X$.} Few top LP-polynomials for $Z$ continuous are given below:
\bea
T_1(z;F_Z)&=&\Leg_1\big( F_Z(Z) \big)\,=\, \sqrt{12}\big( F_Z(z) - 1/2 \big) \label{eq:leg1}\\
T_2(z;F_Z)&=&\Leg_2\big( F_Z(Z) \big)\,=\,\sqrt{5}\big( 6 F^2_Z(Z) - 6 F_Z(Z) + 1 \big) \label{eq:leg2}\\
T_3(z;F_Z)&=&\Leg_3\big( F_Z(Z) \big)\,=\,\sqrt{7}\big( 20 F^3_Z(Z) - 30 F^2_Z(Z) + 12 F_Z(Z) -1    \big). \label{eq:leg3}
\eea
and so on. 
For more information, ses \cite{DD20181,Deep17LPMode1}. This will be useful for constructing LP-polynomials for any general pivot density $f_0$ of the response variable. 
\vskip.15em

4) Derivation of empirical $T_1(z;\wtF_Z)$ from a sample of all unique $z_1,\ldots,z_N$ (i.e, realization from a continuous $Z$). In this case, the empirical probability mass function (pmf) is $\tp_Z(z_i)=1/N$, and the empirical cdf is $\wtF_Z(z_i)=R_i/N$, where $R_i$=\texttt{rank}($z_i$). This immediately implies (following eq. \ref{eq:LP1st})
\beq \label{eq:eLP1st}
T_1(z;\wtF_Z)~=~\sqrt{\frac{12}{N^2-1}}\Big(R_i - \frac{N+1}{2}\Big).
\eeq
Since,
\[ \tFm_Z(z_i)= \wtF_Z(z_i)- \frac{1}{2}\tp_Z(z_i)= \frac{1}{N} (R_i - 1/2).\]
and the correction factor
\[ 1-\sum_i \tp_Z^3(z_i) = 1 - \frac{1}{N^2}. \]
Eq. \eqref{eq:eLP1st} makes it clear why we call LP-bases are polynomials of ranks. 

\vskip.15em
5) The shapes of LP-polynomials are data-adaptive. The left panel of Fig. \ref{fig:app_lppoly} displays the top four \texttt{eLP}-basis functions for the covariate \texttt{weight}, taken from the Auto-MPG data (Example 12 in the main paper). The weight variable has very little ties, which is the reason why the shapes of LP-bases completely match with eq. (\ref{eq:leg1}-\ref{eq:leg3}). Contrast this with the right panel of Fig. \ref{fig:app_lppoly}, which shows top four \texttt{eLP} bases for the variable \texttt{acceleration}. Due to the presence of a large number of ties, it takes a unique piecewise-constant shape.

\vskip1em
\subsection*{A.3.~Remaining Proofs}
\vskip.35em
{\bf Theorem \ref{thm:entropydiff}}: Since the conditional entropy $H(Y|X)=\int H(Y|X=x) \dd F_{X}(x)$, we have
\beas 
H(Y) - H(Y|X) &=& - \int f_Y(y) \log f_Y(y) \dd y + \iint f_{Y|x}(y) \log f_{Y|x}(y) \dd y \dd F_X(x)\\
&=& \int \Big \{ \int f_{Y|x}(y) \log \frac{f_{Y|x}(y)}{f_Y(y)} \dd y \Big\} \dd F_X(x)\\
&=& \Ex_X{\rm KL}(d_X;\mathbbm{1}_{[0,1]}).
\eeas
The last equality follows from Theorem \ref{thm:kl}.
\vskip.35em
{\bf Theorem \ref{thm:kw}}: The Kruskal–Wallis statistic \eqref{eq:kw} is given by (up to a negligible factor $(n-1)/n$)
\beq \label{eq:skw1}\mbox{KW}~=~\sum_{\ell=1}^k n_\ell \Bigg\{ \sqrt{\frac{12}{N^2-1}}\big( n_\ell^{-1}\sum_{i\in G_\ell} R_i - \frac{N+1}{2} \big)  \Bigg\}^2
\eeq
Now note that 
\beq \label{eq:skw2}
\tLP_{1|\ell}~=~\Ex[T_1(Y;\wtF_Y)|X=\ell] ~=~\frac{1}{n_\ell} \sum_{i \in G_\ell} \sqrt{\frac{12}{N^2-1}}\Big(R_i - \frac{N+1}{2}\Big).
\eeq
The first equality follows from \eqref{eq:lpcof1} and the second one from \eqref{eq:eLP1st}. Combining \eqref{eq:skw1} and \eqref{eq:skw2} we get the first desired result
\beq \label{eq:skw3}
\mbox{KW}~=~\sum_{\ell=1}^k n_\ell \,\big| \tLP_{1|\ell} \big|^2.\eeq
Next, to show that \eqref{eq:skw3} is actually equals to $NR_1^2$, note that
\[R_1^2 ~=~\frac{\Var(E[T_1(Y;F_Y)|X])}{\Var(T_1(Y;F_Y))},\]
where the denominator is $1$ by the construction of LP-basis functions, and the numerator is $\sum_{\ell=1}^k \pi_\ell \,\big| \tLP_{1|\ell} \big|^2$, $\pi_\ell$ is $\Pr(X=\ell)=n_\ell/N$. This is because
\[\Ex[\LP_{1|X}]~=~\Ex\left[\Ex[T_1(Y;F_Y)|X]\right]~=~\Ex[T_1(Y;F_Y)]=0.\]

{\bf Theorem \ref{thm:dif}}: Define the LP unit-basis functions as 
\beq \label{eq:Ulp}
S_j(u;F_Y)=T_j(Q_Y(u);F_Y) ~~{\rm for}~ 0<u<1,\eeq
which satisfy the following orthonormality relations (immediate from eq. \ref{eq:onor})
\beq \label{eq:sor} \int_0^1 S_j(u;F_Y) S_k(u;F_Y) \dd u\,=\,\delta_{jk}.\eeq

Substitute $F_Y(y)=u$ in \eqref{eq:dLP} to express the contrast density function in the quantile domain:
\[d_{x,z}(u)~=~1+\sum_{j} \LP_{j|x,z} S_j(u;F_Y),~0<u<1.\]
It is now easy to see that the $L^2$ distance between $d_{x,0}$ and $d_{x,1}$ can be written as
\[\big \| d_{x,0} - d_{x,1} \big\|_2^2 \,=\, \sum_{j} \big| \LP_{j|x,0}\,-\, \LP_{j|x,1}\big|^2 \]
due to \eqref{eq:sor}. Hence proved.
%%%%%%%%%%%%%%%55
\vskip1em
\subsection*{A.4.~Additional Remarks} \label{app:proof}

\begin{figure}[t]
 \centering
 \includegraphics[width=.7\linewidth,trim=1cm .5cm 1cm .5cm]{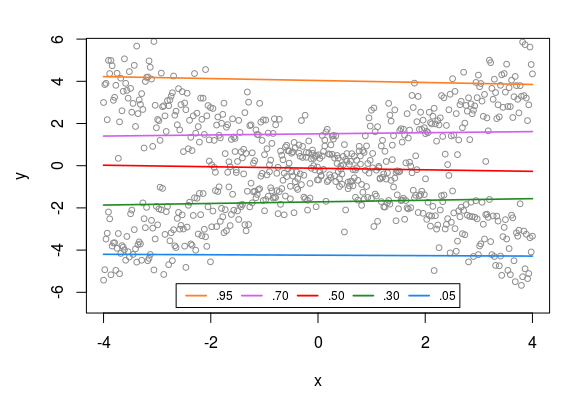}
\caption{Butterfly data: Performs Roger Koenker's (\citeyear{koenker1978q}) quantile regression routine as implemented in \texttt{rq}() function of the R-package \texttt{quantreg}. Note the difference with Fig. \ref{fig:app_quantreg}.} \label{fig:kqr}
\end{figure}

\vskip.35em
{\bf Remark A.4.1}~ \textit{Koenker's Quantile regression for Butterfly data}:
Fig. \ref{fig:kqr} shows the quantile regression curves for the butterfly data, which implements the algorithm proposed in \cite{koenker1987algorithm}.

\begin{figure}[t]
    \centering
    \begin{subfigure}{.48\linewidth}
    \centering
        \includegraphics[width=\linewidth,trim=1cm 0cm 0cm .5cm]{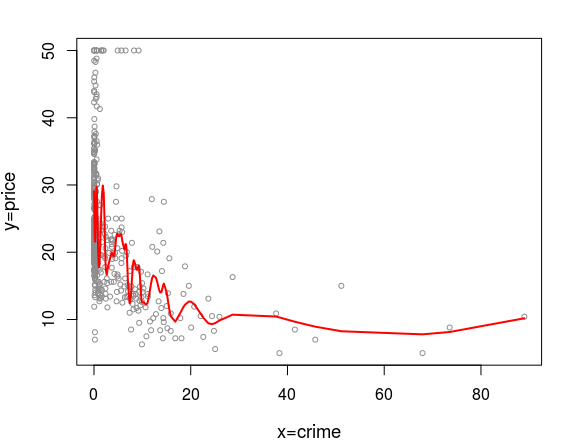}
    \end{subfigure}~~
    \begin{subfigure}{.48\linewidth}
    \centering
        \includegraphics[width=\linewidth,trim=0cm 0cm 1cm .5cm]{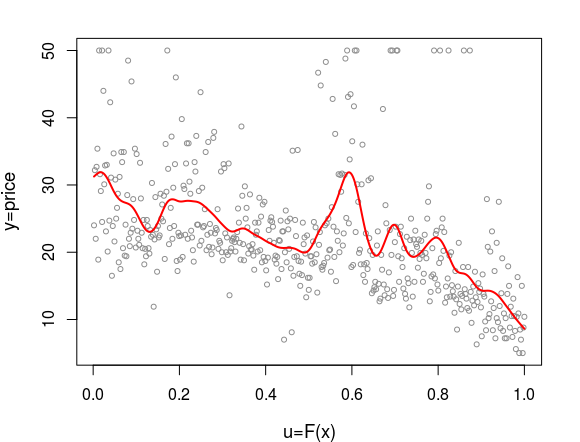}
    \end{subfigure}~~
    \caption{Boston housing data: The house prices drop sharply with high crime rates.}
    \label{fig:app_boston}
\end{figure}

\vskip.35em
{\bf Remark A.4.2}~ \textit{Smoothness of Regression via Rank-transform}. Fig. \ref{fig:app_boston} shows 
house \texttt{prices} (median value of in 1000s) in Boston metropolitan area as function of \texttt{crime} (per capita crime rate). The data were collected in 1970 and available in the UCI Machine Learning Repository. The left plot shows the data in the xy-domain, and the right plot shows the same data in the quantile domain where we replaced $x$ by its rank-transform $\wtF_X(x)$.  Red curves are the spline smoother (estimated using \texttt{smooth.spline}() R-function). Few interesting points to note:

~$\bullet$ Smoothness: Regression \emph{via} rank-transform (see  Theorem \ref{thm:rankreg}) seems to produce much smoother (parsimonious) regression curve compared to the rough zigzag xy-domain estimate. Is it universal that nature reveals herself in a more parsimonious way in the quantile domain? Apart from intuitive understanding, we still don't know any concrete mathematical explanation for this surprising phenomenon.

~$\bullet$ Data-sparsity: The quantile-domain treatment seems to address the `data sparsity' (that exists for $x>30$) problem quite well. Surprisingly, after quantile-transformation, a data-sparse regression problem turns into a data-dense one.

~$\bullet$ Visualization: The rank-transform scatter makes it far easy to visualize and understand the relationship between $x$ and $y$. 

~$\bullet$ Robust+Smooth: This shows an additional benefit of X-robustness (Theorem \ref{thm:rankreg}) as a tool for promoting smoothness.

\begin{figure}[h]
\centering
\begin{subfigure}[t]{.45\linewidth}
    \centering
    \caption{GRF}
        \includegraphics[width=\linewidth,trim=1cm 0cm 0cm .5cm]{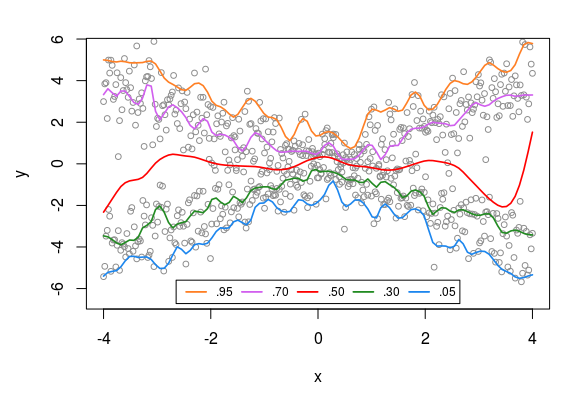}
    \end{subfigure}~~
    \begin{subfigure}[t]{.45\linewidth}
    \centering
    \caption{GBM}
        \includegraphics[width=\linewidth,trim=1cm 0cm 0cm .5cm]{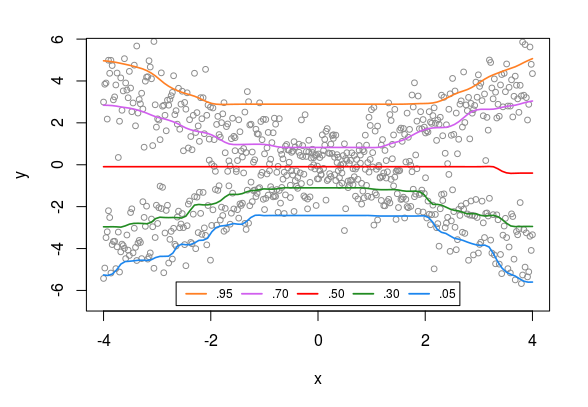}
    \end{subfigure}\\
    \begin{subfigure}[t]{.45\linewidth}
    \centering
    \caption{UPM+RF}
        \includegraphics[width=.9\linewidth,trim=2cm 0cm 0cm .5cm]{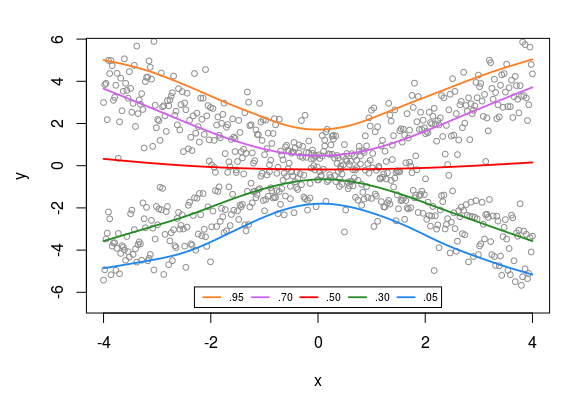}
    \end{subfigure}~~
    \begin{subfigure}[t]{.45\linewidth}
    \centering
    \caption{UPM+GBM}
         \includegraphics[width=\linewidth,trim=1cm 0cm 0cm .5cm]{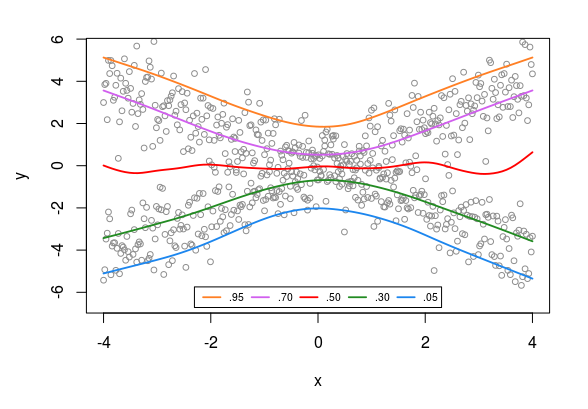}
    \end{subfigure}\vskip-1em
    \caption{Butterfly data: Quantile Regression comparisons. (a) generalized random forests based quantile regression curves (as implemented in the \texttt{grf} R-package), which are too wiggly even after spline smoothing (with default degrees of freedom). (b) Generalized boosted quantile regression curves (as implemented in \texttt{gbm} R-package). They suffer from under-smoothing in the central region and over-smoothing at the tails. The bottom row shows our UPM-versions.} \label{fig:app_quantreg}
        \end{figure}

\vskip.35em
{\bf Remark A.4.3}~ \textit{ML-assisted Quantile regression: comparison}. The first row of Fig. \ref{fig:app_quantreg} shows the generalized random forests-based \citep{athey2019grf1} and gradient boosted quantile regression estimates. They are specially-designed ML-algorithms to produce the desired conditional quantile curves. The second row shows our UPM-based versions, which are derived from a general scheme described in Section \ref{sec:dest} of the main paper.

\begin{figure}[h]
 \centering
 \includegraphics[width=.48\linewidth,trim=1cm .5cm .25cm .5cm]{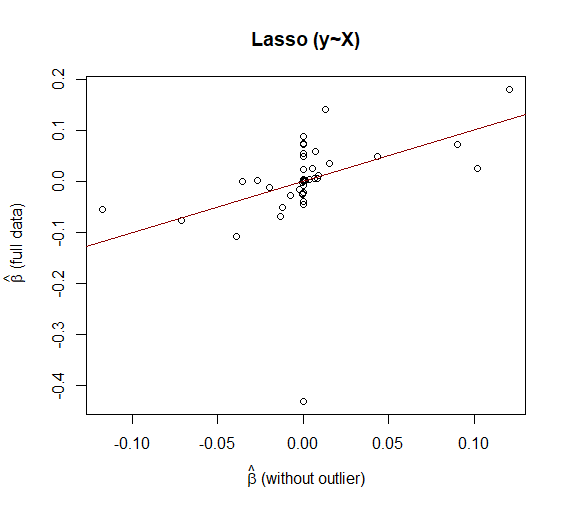}~~~
  \includegraphics[width=.48\linewidth,trim=.25cm .5cm 1cm .5cm]{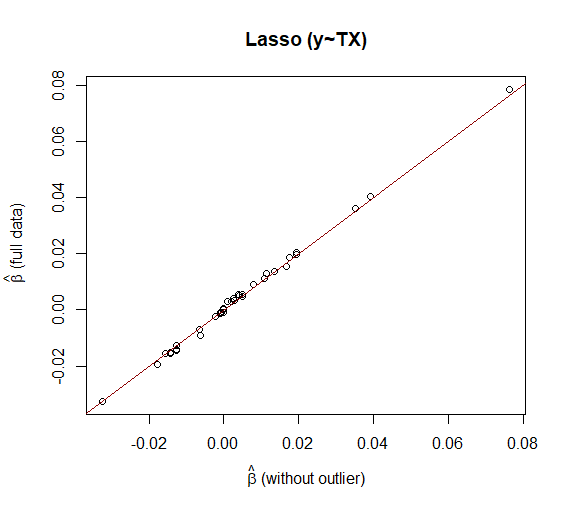}
  \vskip1em
\caption{Online news popularity data. The goal is to assess the impact of a single outlying observation (the news article in row \#$31,038$ on the lasso-estimated coefficients. The left plot shows the result for \texttt{lasso}($y \sim X$) case, and the right one for robustified \texttt{lasso}($y \sim T_X$). For more details see remark A.4.4.} \label{fig:rlasso}
\end{figure}

\vskip.35em
{\bf Remark A.4.4}~ \textit{Robust Lasso}: As already noted, the online news data contain an outlier--the news article in row \#31,038. How does it impact traditional estimation and feature selection methods? As we will see, a single data point can have a devastating impact on the overall modeling. Fig. \ref{fig:rlasso} (left panel) shows the lasso estimated $\widehat{\be}$ under two scenarios: first, on the outlier-removed data, and second, on the full data (with outlier). Notice the estimated coefficient values (as well as their sparsity-levels) are dramatically affected because of just one data point.  One way to address this fragility is to apply lasso on the first-order LP-transformed $T_X=[T_1(x;\wtF_{X_1}),\ldots, T_1(x;\wtF_{X_p})]$, instead of raw feature matrix $X$. The result is displayed on the right panel, which shows extraordinary consistency with (almost) no impact on the estimated coefficients. This mid-rank transform based \emph{robust lasso} method could be a reliable alternative to deal with noisy messy data. 
%%%apendix refs

\newpage

\end{document}